\documentclass{article}
\usepackage{iclr2027_conference,times}

\usepackage{amsmath,amsfonts,bm}

\def\eqref#1{equation~\ref{#1}}

\def\1{\bm{1}}

\DeclareMathAlphabet{\mathsfit}{\encodingdefault}{\sfdefault}{m}{sl}
\SetMathAlphabet{\mathsfit}{bold}{\encodingdefault}{\sfdefault}{bx}{n}

\usepackage{hyperref}
\usepackage{url}
\usepackage{verbatim}
\usepackage{tikz}
\usepackage{xcolor}
\usepackage{wrapfig}
\usepackage{tabularx}
\usepackage{pgfplots}
\pgfplotsset{compat=1.18}
\usepackage{subcaption}
\usepackage{booktabs,multirow}
\usepackage{tikz,pifont}
\usepackage{enumitem}
\usepackage{siunitx}
\usepackage{placeins}
\usepackage{float}
\usepackage{longtable}
\usepackage{booktabs}
\usepackage{array}
\usepackage{enumitem}

\newcommand{\dynamicdxtitle}{DYNAMICDX: Evaluating Evidence\\ Acquisition in Video-Based Diagnosis}
\title{\dynamicdxtitle}

\author{Jiahui Li$^{1}$, Yutong Guo$^{1}$, Nan Yang$^{2}$, Wenzhan Song$^{1}$, Jin Lu$^{1}$, Fei Dou$^{1}$\thanks{Corresponding author.} \\
$^{1}$University of Georgia, Athens, USA \qquad $^{2}$Beijing Luhe Hospital, Beijing, China \\
\texttt{\{jl57095,\,yutong.guo,\,wsong,\,jin.lu,\,fei.dou\}@uga.edu} \\
\texttt{yangnan@mail.ccmu.edu.cn}
}

\iclrfinalcopy

\begin{document}
\newcommand{\partonereference}[7]{%
\begin{tikzpicture}
\node[anchor=south west,inner sep=0] (plot) at (0,0)
  {\includegraphics[width=\linewidth]{#1}};
\begin{scope}[x={(plot.south east)},y={(plot.north west)}]
\pgfmathsetmacro{\referencePct}{100*#2/71}
\pgfmathsetmacro{\referenceY}{(656-545+(545-#4)*(100*#2/71)/#3)/656}
\draw[gray!75!black,line width=0.65pt,dash pattern=on 3pt off 1.4pt]
  (114/900,\referenceY) -- (880/900,\referenceY);
\node[anchor=south west,inner sep=0.7pt,text=gray!75!black,
  font=\sffamily\fontsize{5.5}{6}\selectfont]
  at (#7,\referenceY+0.009)
  {clinician (\pgfmathprintnumber[fixed,precision=1,zerofill]{\referencePct}\%)};
\fill[white] (#5,#6-0.025) rectangle (#5+0.207,#6+0.025);
\node[anchor=west,inner sep=0,font=\sffamily\fontsize{5.5}{6}\selectfont]
  at (#5,#6) {qwen3.8-flash};
\end{scope}
\end{tikzpicture}%
}

\maketitle
\lhead{Preprint}

\begin{abstract}
Diagnosing a patient from video requires more than recognizing the sign: a
vision--language model must turn what it sees into hypotheses, questions and tests.
\textsc{DynamicDx} evaluates each step in 71 neurological consultations across
11 sign categories, linking authentic patient videos to confirmed
diagnoses and fixed charts built from the same case reports, so that every
model queries the same evidence. Across five such
models, video improves accuracy by 9.9--22.5 percentage points over blind
input, but neither recognition alone nor temporal order explains the gain: the cause
is usually missing from the model's video-only differential diagnosis even when the sign is
recognized, and shuffling the frames produces no reliable accuracy loss.
Instead, a trajectory replay traces most of the gain to the investigation
results the video prompts. Evidence acquisition is the bottleneck: supplying
the decisive investigations raises accuracy to 73.2--93.0\%.
Two interventions act on it. A post-trained 4B video describer improves sign
descriptions, especially from a short, densely sampled segment, and source-clean
literature retrieval expands initial hypotheses; both bring the tests a model orders
closer to those the treating clinicians documented and, through them, raise accuracy.
For video-based diagnosis, seeing better helps when it leads to asking better.
\end{abstract}

\section{Introduction}
\suppressfloats[t]

\begin{figure}[t]
\centering
\includegraphics[width=\textwidth]{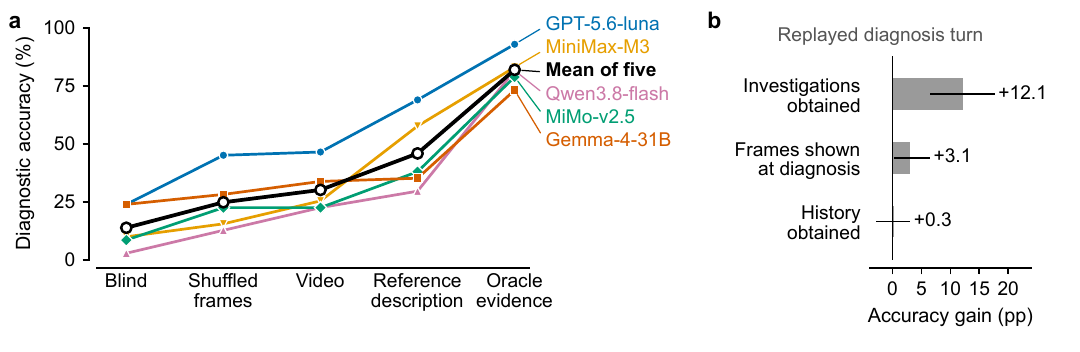}
\caption{Video helps diagnosis through the evidence it prompts, not through its dynamics.
(a)~Diagnostic accuracy on the 71 cases as the input grows (Table~\ref{tab:consultation_conditions}); black, mean over the five models.
(b)~Accuracy gained at a replayed diagnosis turn from each component, mean over the five models with $95\%$ bootstrap intervals (Table~\ref{tab:replay}).}
\label{fig:headline}
\end{figure}

Clinical diagnosis links presenting signs to hypotheses, evidence acquisition,
and a final decision. In neurology, a sign's phenomenology, what the movement
looks like, narrows its plausible causes (aetiologies) and the investigations to
order; irregular flowing movements (chorea) and sustained postures (dystonia),
for example, point to different disorders
\citep{abdo2010clinical,bhatia2018consensus,albanese2013phenomenology}.
Recognizing a sign is only the beginning: a model must list candidate causes
(the differential diagnosis), question the patient (history-taking), order tests
(the work-up), and integrate the returned evidence. A correct description can
therefore coexist with an incomplete differential or an incorrect diagnosis.
Evaluating only the final answer hides these steps.

Existing benchmarks capture parts of this process. Text-based systems
assess differential diagnosis and patient interaction
\citep{mcduff2025towards,tu2025towards}, while multimodal evaluations use
static images, procedural videos, or actor-based encounters
\citep{gupta2023dataset,schmidgall2026agentclinic,shah2026towards}.
Two questions remain open. Do models use how a patient's sign unfolds over time, not only how it looks? And how does what they see shape the evidence they acquire?

We introduce \textsc{DynamicDx}, a process-level diagnostic environment driven
by real patient videos: 71 clips across 11 neurological phenomenological
categories, each paired with the confirmed diagnosis and a fixed chart from the
same source report. Consultations proceed in two stages, so that a missed
diagnosis can be traced to what the model saw or to the evidence it sought:
Stage~1 measures reactive perception (recognizing the sign and proposing a
differential), and Stage~2 proactive evidence acquisition (taking a history and
ordering investigations) before the diagnosis. Every model queries the same fixed chart,
so differences in outcome reflect the model rather than a simulated patient.
Charts are semi-synthetic: expected values fill in the test results a report
left out, so the chart is complete and which tests return results does not
reveal the diagnosis. Seven controlled conditions vary visual input, textual
representation, and supplied evidence; single-frame and shuffled inputs separate
additional views from their temporal order.

Across five vision--language models, video improves accuracy by
9.9--22.5 percentage points over blind input (Figure~\ref{fig:headline}). On the
first question, current models use what a clip shows rather than how it unfolds:
shuffling the frames produces no reliable accuracy loss, a gap that single-answer
video benchmarks \citep{fu2025video} do not expose. On the second, what they see
helps through the evidence it prompts rather than through recognition: recognizing
the sign rarely puts its cause in the differential, and replaying consultations
with swapped trajectories traces most of the gain at the diagnosis step to the
investigation results obtained, not to the history. Evidence
acquisition is the bottleneck: clinician-written descriptions recover more of
the documented work-up and raise accuracy, and supplying the decisive
investigations raises it to 73.2--93.0\%. Two interventions act on it: a
post-trained video describer improves sign descriptions, especially from a short,
densely sampled segment, and source-clean retrieval expands hypotheses; both raise
accuracy by moving the work-up toward the documented one.
Our contributions are:
\begin{itemize}
    \item \textbf{A benchmark for evidence acquisition from patient video.}
    \textsc{DynamicDx} builds interactive consultations on authentic patient
    video, pairing each clip with the confirmed diagnosis and a fixed chart from
    the same case report, so that models are compared on identical evidence and a
    missed diagnosis can be traced to the stage where it failed.
    \item \textbf{Evidence acquisition as the bottleneck.} We show that video
    helps diagnosis through the investigations it prompts rather than through
    recognition or temporal order, and that accuracy is limited by acquiring the
    decisive investigations rather than by interpreting them.
    \item \textbf{Two ways to act on it.} A post-trained video describer and
    source-clean retrieval each raise accuracy by moving the work-up toward the
    documented one, and controls separate sampling density from window choice and
    list breadth from retrieved content.
\end{itemize}

\section{Related Work}

\paragraph{Sequential clinical reasoning and conversational diagnosis.}

Clinical reasoning links presenting findings to diagnostic hypotheses and
evidence acquisition \citep{elstein1978medical,elstein2002clinical,croskerry2009universal}.
Medical VQA and question answering evaluate answers from fixed inputs
\citep{lau2018dataset,singhal2023large,singhal2025toward,moor2023foundation,saab2024capabilities,jeong2024medical}, whereas interactive systems assess question asking,
diagnostic dialogue, multi-agent deliberation, and tool use
\citep{li2024mediq,johri2024craft,johri2025evaluation,tu2025towards,kim2024mdagents,schmidgall2026agentclinic,lievin2026towards}; selective evidence acquisition also improves clinical signal classification \citep{li2026deeparrhythmia}.
Recent work also evaluates differential diagnosis from case descriptions
\citep{mcduff2025towards}, multimodal artifacts \citep{saab2026advancing}, and
live video consultations with patient actors \mbox{\citep{shah2026towards}}.
The interactive multimodal evaluations in Table~\ref{tab:benchmark-comparison} span $20$--$120$ cases or scenarios, none drawn from patient video;
\textsc{DynamicDx} instead links $71$ authentic patient clips to fixed source-case
history and test results and scores each stage from recognition to diagnosis.

\begin{table}[t]
\centering
\caption{Benchmark comparison (\ding{51}: included; Diff.: differential; Hx: history; Dx: diagnosis). Cases: evaluation cases or scenarios (MedVidQA: videos; AgentClinic: multimodal + text).}
\label{tab:benchmark-comparison}
\footnotesize
\setlength{\tabcolsep}{.5pt}
\renewcommand{\arraystretch}{1.12}
\begin{tabularx}{\linewidth}{@{}>{\raggedright\arraybackslash}p{0.22\linewidth}>{\raggedright\arraybackslash}p{0.23\linewidth}>{\raggedleft\arraybackslash}p{0.09\linewidth}*{6}{>{\centering\arraybackslash}X}@{}}
\toprule
Benchmark & Source & Cases & Video & Sign & Diff. & Hx & Tests & Dx \\
\midrule
MedVidQA
  & Instructional video & 899 & \ding{51} & -- & -- & -- & -- & -- \\
MediQ
  & Simulated cases & 1{,}413 & -- & -- & -- & \ding{51} & -- & \ding{51} \\
AgentClinic
  & Simulated cases/records & 120\,+\,415 & -- & -- & -- & \ding{51} & \ding{51} & \ding{51} \\
Multimodal AMIE
  & Actors + artifacts & 105 & -- & -- & -- & \ding{51} & -- & \ding{51} \\
AI co-clinician
  & Actors & 20 & \ding{51} & -- & -- & \ding{51} & -- & \ding{51} \\
\midrule
\textbf{DynamicDx (ours)}
  & \textbf{Patient clips + records} & \textbf{71} & \ding{51} & \ding{51} & \ding{51} & \ding{51} & \ding{51} & \ding{51} \\
\bottomrule
\end{tabularx}
\end{table}

\paragraph{Dynamic medical video understanding and temporal sampling.}
MedVidQA evaluates medical-video classification and temporal answer localization
\citep{gupta2022overview}; video analysis supports tremor, Parkinson's disease, dystonia and seizure assessment \citep{friedrich2024validation,di2025ai,haberfehlner2023towards,ahmedt2024deep} and remote movement-disorder examination \citep{srinivasan2020telemedicine,newton2022through,angelopoulou2024neurological}. In general-domain video, many questions can be answered from a single frame and temporal understanding remains weak \citep{buch2022revisiting,liu2024tempcompass}, echoing the reliance on priors in image VQA \citep{goyal2017making,agrawal2018don}. General video sampling methods use sequential
selection \citep{wu2019adaframe}, relevance--coverage optimization
\citep{tang2025adaptive}, adaptive selectors \citep{buch2025flexible}, or
generative relevance supervision \citep{yao2025generative} to address redundancy
and context limits. Clinical clips present a complementary challenge: brief or
high-frequency signs may be poorly captured by uniform sampling, and instruction tuning can adapt language models to physiological signals \citep{li2026peakdetector}. We evaluate
temporal-window selection through its effects on downstream evidence acquisition
and diagnosis, beyond video recognition alone.

\paragraph{Retrieval-augmented clinical reasoning.}
Clinicians routinely consult the literature \citep{westbrook2004clinicians,weller2023information,weller2026systematic}.
MedRAG benchmarks retrieval configurations for medical QA
\citep{xiong2024benchmarking}, i-MedRAG supports iterative information seeking
\citep{xiong2025improving}, and RULE examines retrieved-context reliability in
multimodal QA \citep{xia2024rule}. In \textsc{DynamicDx}, retrieval supplies
candidate aetiologies from predicted phenomenology before history-taking.

\section{Method}

\subsection{Benchmark Construction and Case Representation}
\label{sec:dataset}

\textsc{DynamicDx} contains 71 consultations from 66 open-access case reports
with patient videos, spanning 11 neurological phenomenological categories:
chorea (10), paroxysmal events (9), central vertigo (9), functional movement
disorder (7), ataxia (6), dystonia (6), parkinsonism (6), ptosis (6), tremor
(6), myoclonus (5), and facial palsy (1). Reports retrieved from Europe
PMC~\citep{europe2015europe} were retained when the video visibly demonstrated
the presenting sign and the report documented a confirmed diagnosis and
workup. Videos were trimmed to at most 30 seconds around the sign; findings
not established visually were represented as history or investigation
evidence. Sources and licences are listed in Appendix~\ref{app:attribution}.

Each case is $c=(V,H,E,D,T)$: video $V$, documented positive and negative
history findings $H$, investigation results $E$, diagnosis $D$, and decisive
entries $T\subseteq E$ that support the diagnosis or exclude an alternative.
Free-form questions and orders map to $H$ and $E$. A case report lists only the
tests its clinicians ran, so a chart built from it alone would leave reasonable
tests unanswered, even when other reports establish their results for the
condition. It would also give the answer away: any test that returned a result
would point toward the diagnosis. $E$ is therefore semi-synthetic. All cases of a category share one menu of tests;
a test the report ran returns its reported result, and every other test returns
a derived value expected for the patient's condition, usually
normal or not performed. Derived values were set by annotators or by
DeepSeek-V4.1-Flash given the confirmed diagnosis, and fixed before evaluation
(Appendix~\ref{app:records}). A second, blinded annotator independently labeled decisive
entries (Cohen's $\kappa=0.75$), with disagreements resolved by consensus.

\subsection{Sequential Diagnostic Environment}
\label{sec:diagnostic_environment}

\begin{figure}[t]
    \centering
    \includegraphics[
        width=1\textwidth,
        trim=10 0 0 0,
        clip
    ]{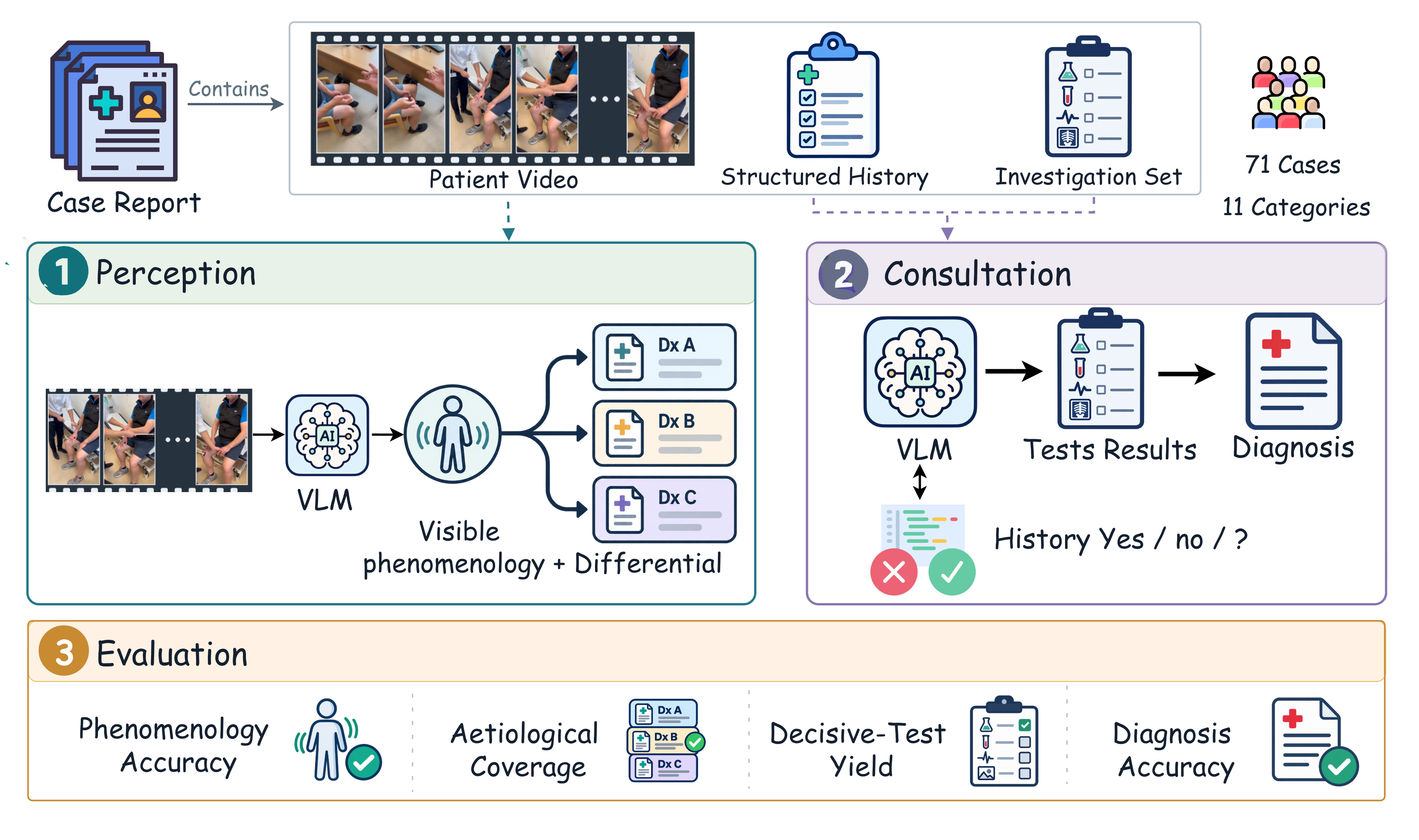}
    \caption{Overview of \textsc{DynamicDx}. Stage~1 evaluates reactive perception (sign
    recognition and hypotheses from video); Stage~2 evaluates proactive evidence
    acquisition (history and investigations) and diagnosis from the case chart.}
    \label{fig:dynamicdx_framework}
\end{figure}

Consultations proceed in two stages (Figure~\ref{fig:dynamicdx_framework}), so that a missed
diagnosis can be traced to what the model saw or to the evidence it sought.
Stage~1 measures reactive perception: from $K$ temporally ordered frames alone,
the model describes the phenomenology and gives an open-ended differential.
Stage~2 measures proactive evidence acquisition: the model asks all its history
questions in one turn and orders all investigations in a second, then gives its diagnosis.
By default, Stage~2 starts from the video alone; own-words and retrieval conditions supply Stage~1 output instead.

Fixed prompts on DeepSeek-V4.1-Flash \citep{deepseek2026v41flash} map free-text questions and
orders to the chart, so that every model is answered from the same fixed record (Appendix~\ref{app:prompts}). A history question is answered \texttt{yes} or \texttt{no} only when the record documents
presence or absence, and \texttt{unknown} otherwise.
An order releases the fixed results of the entries it matches, and unmatched orders return \texttt{not performed / not available}. Orders may name therapeutic trials, such as a levodopa challenge. The model gives one primary diagnosis
and up to three alternatives; only the primary diagnosis is scored (Section~\ref{sec:measurements}).

\subsection{Process-Level Measurements}
\label{sec:measurements}

We evaluate four outcomes. \textbf{Phenomenology recognition} credits the
reference visible sign, written independently by two annotators from reports and
videos ($\kappa=0.83$), or equivalent terminology, distinguishing partial
from incorrect descriptions. \textbf{Aetiological coverage} considers the
full video-only differential, including entries beyond the first three;
compatible syndromes or broader aetiological categories qualify without
requiring the confirmed disease. \textbf{Decisive test yield} asks whether a
consultation obtains the results that settled the case. It is measured by
source-workup coverage, $\tau=|T\cap E_{\mathrm{acq}}|/|T|$, the share of a case's
decisive entries $T$ found among the chart entries $E_{\mathrm{acq}}$ a consultation
obtains, pooled over the $71$ cases; an entry reporting the same finding from the
same kind of test also counts (Appendix~\ref{app:tau}). $\tau$ measures how
much of the documented work-up is recovered. \textbf{Final diagnosis} uses case-specific
acceptance lists: \textsc{Accurate} includes accepted disease or syndrome
formulations; \textsc{Partial} meets partial-credit criteria or identifies
only the broad disease family; \textsc{Not accurate} meets neither criterion.
Clinical experts independently built the acceptance lists, blinded to all model outputs.
Diagnostic accuracy is the proportion graded \textsc{Accurate}.

All outcomes are graded automatically by separate DeepSeek-V4.1-Flash prompts, given with
the grading criteria in Appendix~\ref{app:prompts}. Independent human review agrees with the
automated grades on $89.6\%$ of phenomenology and $96.9\%$ of aetiology assessments
($96$ Stage~1 responses) and on $91.5\%$ of diagnostic grades for all $71$ GPT-5.6-luna
video consultations ($\kappa_w=0.947$). Of $100$ random orders across five models,
$95$ match correctly, with three over-releases and two misses (Appendix~\ref{app:grader}).

\subsection{Controlled Diagnostic Conditions}
\label{sec:controlled_conditions}

Seven conditions locate where the benefit of video arises; each changes one input and keeps
the rest of the consultation fixed. \textbf{Video} provides $K=32$ temporally ordered frames
throughout the consultation, and \textbf{Blind} removes them, stating that the model has not
yet seen or examined the patient; their difference is the total benefit of visual access.
\textbf{Single frame} keeps one frame, testing whether multiple views matter, and
\textbf{Shuffled} presents the same frames in randomly permuted order (three permutations per
case), testing whether temporal order matters. \textbf{Own words} replaces the video with the
model's own phenomenology description, testing whether the clip carries anything the model
cannot put into words, and \textbf{Reference} substitutes a clinician-written description
restricted to visible findings, testing whether better perception would help.
\textbf{Oracle evidence} keeps the Video condition's frames and history but replaces
investigation selection with the adjudicated decisive investigations and their results, so
$E_{\mathrm{acq}}=T$ and $\tau=1$; it measures how well models diagnose once evidence acquisition is solved.

\subsection{Targeted Interventions}
\label{sec:targeted_interventions}

Diagnosis can fail before any evidence is sought, if the sign is misdescribed or a correct
description does not suggest its cause. Temporal-window training (Section~\ref{sec:temporal_training})
targets the first step, describing the phenomenology, and literature retrieval
(Section~\ref{sec:retrieval}) the second, turning it into a differential.

\subsubsection{Teacher-Guided Temporal-Window Training}
\label{sec:temporal_training}

\begin{figure}[t]
    \centering
    \includegraphics[width=\textwidth]{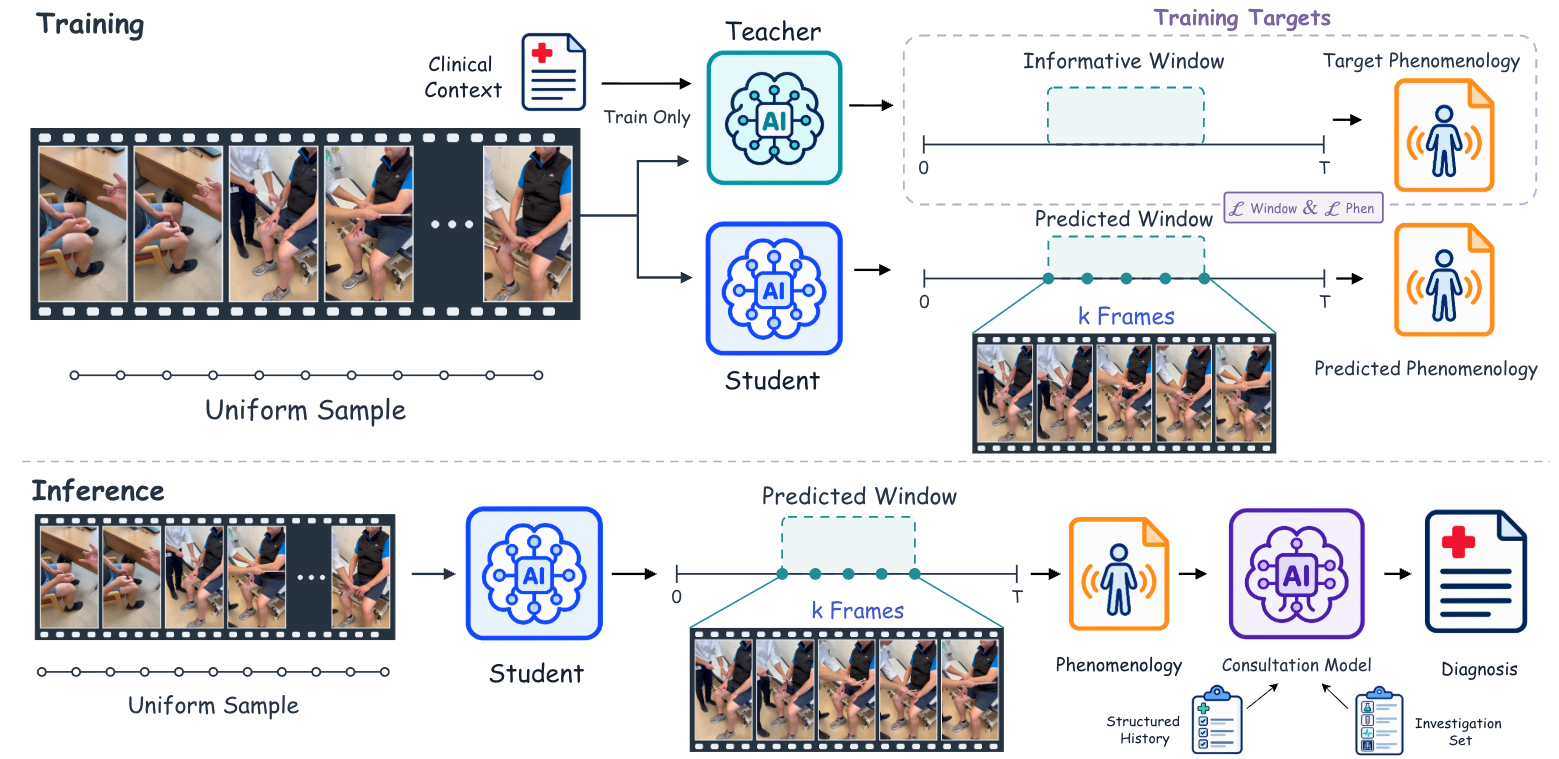}
    \caption{
    Teacher-guided temporal-window sampling. Clinical context supervises
    the offline teacher only; benchmark phenomenology supplies the recognition
    target. The student predicts at most one window from 32 survey frames
    and describes $K$ window frames.
    }

    \label{fig:teacher-guided-windows}
\end{figure}

A dynamic sign often occupies a short segment of a clip, which uniform sampling at small $K$ can
miss, so the describer first chooses where to look. An offline teacher (Figure~\ref{fig:teacher-guided-windows}) uses video, reference phenomenology, and diagnosis in
one call per clip to label at most one window by start time
and duration, or choose no window. We train Qwen3.5-4B \citep{qwen2026qwen35} with LoRA \citep{hu2022lora} to predict
this window from 32 survey frames and describe $K\in\{8,16,32\}$ window
frames, with reference phenomenology as the recognition target.
At evaluation, Window uses the predicted position; Random holds the adapter,
survey, duration, and $K$ fixed but randomizes position; responses without a valid window fall back to the survey.
A sign-only adapter, trained and tested on $32{+}K$ frames spread over the clip, uses no window.
Five-fold cross-validation groups cases by source article. Out-of-fold descriptions initialize
GPT-5.6-luna consultations. Teacher outputs and clinical fields only build training targets; at test time the student sees frames alone.
Appendix~\ref{app:temporal_training_details} gives settings, including the sign-only adapter's one-epoch schedule.

\subsubsection{Literature Retrieval and Decontamination}
\label{sec:retrieval}

\begin{figure}[t]
\centering
\includegraphics[width=\linewidth]{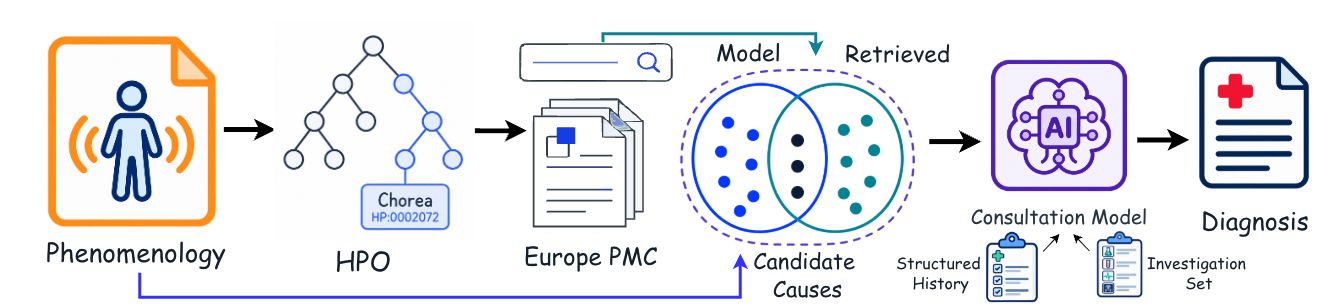}
\caption{Retrieval pipeline from predicted phenomenology to candidate causes.}
\label{fig:retrieval-pipeline}
\end{figure}

DeepSeek-V4.1-Flash normalizes the predicted phenomenology (Figure~\ref{fig:retrieval-pipeline}) to Human Phenotype Ontology
terms \citep{gargano2024human}, which query Europe PMC, and extracts candidate causes from the retrieved
titles and abstract excerpts; these are appended to the model's video-only differential before
history-taking. Queries use the predicted sign, not the confirmed diagnosis. Against the standard
consultation without retrieval, three retrieval conditions test whether gains come from leaked
answers. \textbf{Original} keeps every record; the primary \textbf{Source-clean} condition removes
the source article, shared-DOI records and near-duplicate titles (token Jaccard similarity
$\geq 0.85$); and \textbf{Strict no-answer} also removes records whose titles or excerpts name the
confirmed diagnosis or an accepted equivalent. Two controls test whether gains come from the
retrieved content or from the list itself: \textbf{Own candidates} supplies the model's Stage~1
differential alone, and \textbf{Mismatched retrieval} appends source-clean causes retrieved for a
case from another category, truncated to the same length, so that the list matches in length and
form but not in content. Outcomes are candidate coverage of the true cause, source-workup coverage,
investigation counts and accuracy; Appendix~\ref{app:lit} also audits retained records for same-patient reports.

\subsection{Models and Statistical Analysis}
\label{sec:experimental_setup}

We evaluate Gemma-4-31B \citep{gemmateam2026gemma4technicalreport}, MiMo-v2.5 \citep{xiaomi2026mimov25}, MiniMax-M3 \citep{minimax2026m3}, GPT-5.6-luna \citep{openai2026gpt56}, and
Qwen3.8-flash \citep{qwen2026qwen38flash} through a common serving infrastructure, with a fixed endpoint
per model and temperature~0. Stage~1 uses eight uniformly sampled budgets,
$K\in\{1,2,4,8,16,32,64,128\}$; standard Stage~2 uses $K=32$, with temporal
training budgets specified in Section~\ref{sec:temporal_training}.
We report 95\% percentile intervals from 10,000 case-level bootstrap
resamples. Between-condition comparisons resample the same $71$ cases under both
conditions, reporting percentage-point differences and whether their
intervals exclude zero. 
Appendix~\ref{app:lie} reports a stress test with
corrupted history responses.

\section{Results}
\label{sec:results}

\subsection{Visual Recognition and Hypothesis Formation}
\label{sec:visual_results}

\begin{figure}[tbp]
  \centering
  \includegraphics[width=\linewidth]{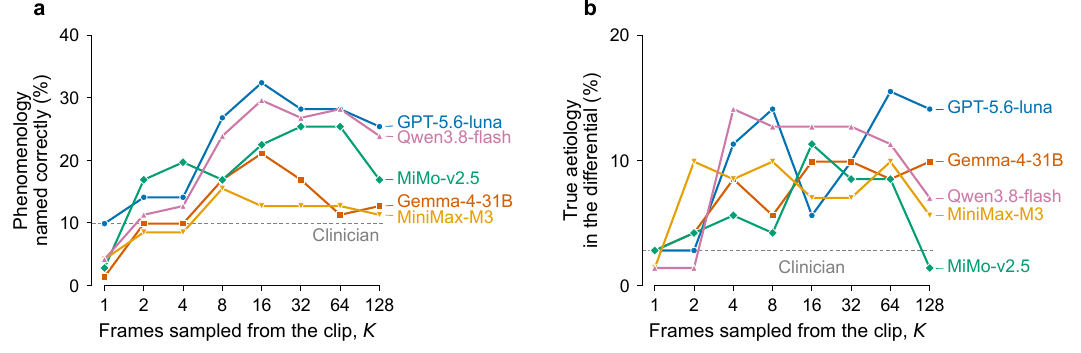}
  \caption{Stage~1 performance on 71 clips. (a)~Phenomenology recognition. (b)~True
  aetiology in the differential. Model curves vary the sampled-frame budget; dashed lines show one
  neurologist's clip-only reference (9.9\% recognition; 2.8\% aetiological coverage).}
  \label{fig:part1}
\end{figure}

Stage~1 shows that correctly describing a movement rarely places its cause in the initial differential. Recognition rises unevenly with the frame budget and peaks at $15.5$--$32.4\%$, depending on the model (Figure~\ref{fig:part1}); at each model's best budget for aetiological coverage, the true cause is still absent from at least $84\%$ of video-only differentials. Even conditional on correct phenomenology, coverage reaches only $18.9\%$ overall ($13.5$--$22.5\%$ across models). Recognizing the sign and forming a useful aetiological hypothesis are thus distinct hurdles (Appendix~\ref{app:frames}), which leaves open whether video helps later, through the questions and tests it prompts.

The clip-only clinician reference underscores both the difficulty of this step and the effect of differential breadth. One neurologist recognized the phenomenology in $7/71$ clips ($9.9\%$), judged $22$ clips unidentifiable, and included the correct aetiology in $2/71$ differentials ($2.8\%$), listing $2.3$ diagnoses per clip against $5.1$--$6.4$ for the models; most model configurations exceed both rates.

\subsection{Diagnostic Outcomes under Controlled Conditions}
\label{sec:consultation_results}
Video improves diagnosis by turning what the model sees into better investigations. Across the five models, video raises accuracy over blind input by $9.9$--$22.5$ percentage points, and source-workup coverage rises with it (Table~\ref{tab:consultation_conditions}): the two models with the largest accuracy gains also gain most in coverage, from $19.3\%$ to $28.9\%$ for GPT-5.6-luna and from $2.0\%$ to $16.3\%$ for Qwen3.8-flash, whereas Gemma-4-31B, the only model whose coverage falls ($22.3\%$ to $20.3\%$), gains least in accuracy. A replay of the diagnosis turn with swapped trajectories confirms this route (Figure~\ref{fig:headline}b; Table~\ref{tab:replay}): swapping in the blind consultation's investigations costs $12.1$ points, swapping its history $0.3$ (the records settle few of the questions asked), and showing the frames at diagnosis adds $3.1$. Part of the benefit runs through derived chart values: neutralising them narrows the video--blind gap from $14.9$ to $10.1$ points (Appendix~\ref{app:tau}). The benefit of video is thus realised in the work-up it prompts rather than in the model's description of the sign.

\begin{table}[t]
\centering
\caption{
Consultation outcomes (\%; $95\%$ bootstrap intervals in brackets).
Seven primary conditions are defined in Section~\ref{sec:controlled_conditions};
multi-turn uses up to ten adaptive rounds. The clinician completed only the video condition (---: not run).
}

\label{tab:consultation_conditions}

\resizebox{\linewidth}{!}{%
\begin{tabular}{llrrrrrr}
\toprule
& & GPT-5.6-luna & Gemma-4-31B & MiMo-v2.5 & MiniMax-M3 & Qwen3.8-flash & Clinician \\
\midrule

\multirow{8}{*}{\shortstack[l]{Diagnosis\\Accuracy (\%)}}
  & Blind
  & 23.9 {\scriptsize [14.1,33.8]} & 23.9 {\scriptsize [14.1,33.8]} & 8.5 {\scriptsize [2.8,15.5]}
  & 9.9 {\scriptsize [4.2,16.9]} & 2.8 {\scriptsize [0.0,7.0]} & --- \\[2pt]
  & Single Frame
  & 36.6 {\scriptsize [25.4,47.9]} & 28.2 {\scriptsize [18.3,39.4]} & 21.1 {\scriptsize [12.7,31.0]}
  & 15.5 {\scriptsize [7.0,23.9]} & 19.7 {\scriptsize [11.3,29.6]} & --- \\[2pt]
  & Shuffled 
  & 45.1 {\scriptsize [33.8,56.3]} & 28.2 {\scriptsize [18.3,39.4]} & 22.5 {\scriptsize [12.7,32.4]}
  & 15.5 {\scriptsize [7.0,23.9]} & 12.7 {\scriptsize [5.6,21.1]} & --- \\[2pt]
  & Video
& 46.5 {\scriptsize [35.2,57.7]} & 33.8 {\scriptsize [22.5,45.1]} & 22.5 {\scriptsize [12.7,32.4]}
& 25.4 {\scriptsize [15.5,35.2]} & 22.5 {\scriptsize [12.7,32.4]} & 35.2 {\scriptsize [23.9,46.5]} \\[2pt]
  & Multi-Turn
  & 52.1 {\scriptsize [40.8,63.4]} & 26.8 {\scriptsize [16.9,36.6]} & 19.7 {\scriptsize [11.3,29.6]}
  & 33.8 {\scriptsize [22.5,45.1]} & 23.9 {\scriptsize [14.1,33.8]} & --- \\[2pt]
  & Own Words
  & 47.9 {\scriptsize [36.6,59.2]} & 33.8 {\scriptsize [22.5,45.1]} & 22.5 {\scriptsize [12.7,32.4]}
  & 22.5 {\scriptsize [14.1,32.4]} & 21.1 {\scriptsize [12.7,31.0]} & --- \\[2pt]
  & Reference
  & 69.0 {\scriptsize [57.7,78.9]} & 35.2 {\scriptsize [23.9,46.5]} & 38.0 {\scriptsize [26.8,49.3]}
  & 57.7 {\scriptsize [45.1,69.0]} & 29.6 {\scriptsize [19.7,40.8]} & --- \\[2pt]
  & Oracle Evidence
  & \textbf{93.0} {\scriptsize [85.9,98.6]} & \textbf{73.2} {\scriptsize [62.0,83.1]}
  & \textbf{78.9} {\scriptsize [69.0,87.3]} & \textbf{83.1} {\scriptsize [74.6,91.5]}
  & \textbf{81.7} {\scriptsize [71.8,90.1]} & --- \\

\midrule

\multirow{7}{*}{\shortstack[l]{Source-Workup\\Coverage (\%)}}
  & Blind
   & 19.3 {\scriptsize [13.1,26.1]} & 22.3 {\scriptsize [15.3,29.4]} & 8.3 {\scriptsize [4.2,13.1]}
   & 13.6 {\scriptsize [8.9,18.8]} & 2.0 {\scriptsize [0.3,4.0]} & --- \\[2pt]
  & Single Frame
   & 28.6 {\scriptsize [22.3,35.4]} & 17.3 {\scriptsize [11.6,23.6]} & 16.9 {\scriptsize [11.1,23.4]}
   & 11.0 {\scriptsize [7.2,15.4]} & 16.3 {\scriptsize [10.3,22.8]} & --- \\[2pt]
  & Shuffled 
   & 32.9 {\scriptsize [27.0,39.2]} & 15.0 {\scriptsize [10.8,19.5]} & 18.9 {\scriptsize [13.5,24.7]}
   & 15.6 {\scriptsize [11.0,20.9]} & 13.6 {\scriptsize [9.3,18.3]} & --- \\[2pt]
  & Video
   & 28.9 {\scriptsize [22.1,35.9]} & 20.3 {\scriptsize [14.9,26.2]} & 16.3 {\scriptsize [11.4,21.5]}
   & 19.3 {\scriptsize [14.8,24.2]} & 16.3 {\scriptsize [11.2,21.9]} & 18.5 {\scriptsize [13.1,24.7]} \\[2pt]
  & Multi-Turn
   & 39.2 {\scriptsize [32.5,45.9]} & 17.6 {\scriptsize [12.5,23.2]} & 14.0 {\scriptsize [9.1,19.2]}
   & 21.6 {\scriptsize [15.6,28.3]} & 10.6 {\scriptsize [6.8,14.8]} & --- \\[2pt]
  & Own Words
   & 39.2 {\scriptsize [32.7,45.8]} & 18.6 {\scriptsize [12.9,24.8]} & 15.3 {\scriptsize [10.7,20.3]}
   & 18.3 {\scriptsize [13.7,23.2]} & 15.3 {\scriptsize [11.6,19.1]} & --- \\[2pt]
  & Reference
   & 50.8 {\scriptsize [43.7,57.9]} & 19.9 {\scriptsize [15.3,24.9]} & 22.3 {\scriptsize [17.5,27.2]}
   & 30.6 {\scriptsize [25.1,36.3]} & 17.9 {\scriptsize [13.4,22.6]} & --- \\

\bottomrule
\end{tabular}%
}
\end{table}

\begin{figure}[t]
    \centering
    \includegraphics[width=\linewidth]{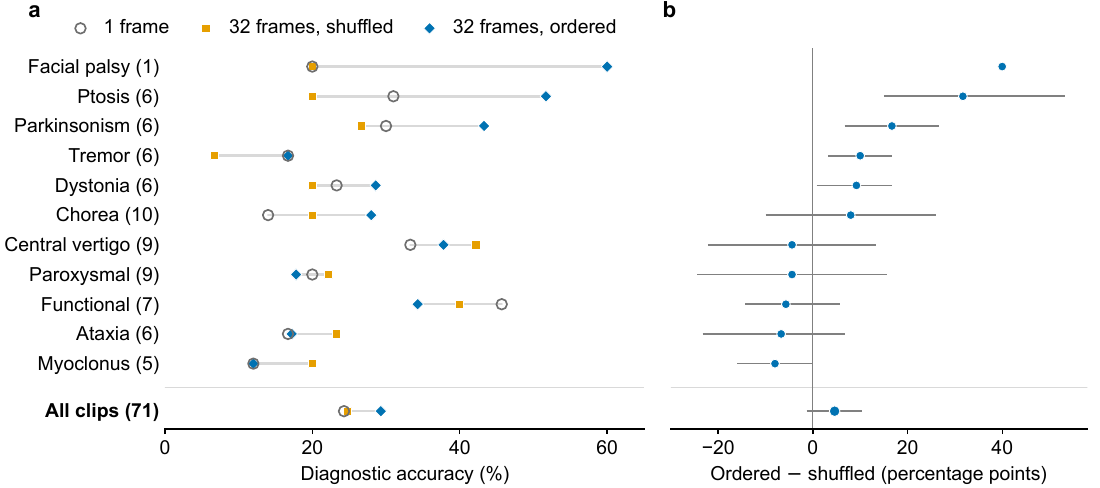}
    \caption{Frame-order effects across clinical categories.
    Five systems pooled over $71$ clips; clip counts in parentheses. (a)~Accuracy for one frame,
    $32$ shuffled frames, and $32$ ordered frames. (b)~Ordered-minus-shuffled
    accuracy with bootstrap intervals ($10{,}000$ clip resamples).}
    \label{fig:frameorder-by-line}
\end{figure}

Better evidence helps most. Clinician descriptions raise accuracy over video by $1.4$--$32.4$ points and raise coverage most where accuracy rises most, and supplying the decisive investigations raises accuracy to $73.2$--$93.0\%$ ($+39.4$--$59.2$ points): correct interpretation of the sign and acquisition of the right evidence both improve diagnosis, acquisition more. By contrast, video, shuffled frames, multi-turn consultation and the model's own words perform alike, consistent with models that read the clip once, largely ignore frame order, and rely on what they can put into words. The model's own description changes accuracy by $-2.8$ to $+1.4$ points relative to video; ordered video exceeds $32$ shuffled frames by a pooled $5.4$ points ($[-0.3, +11.3]$; Figure~\ref{fig:frameorder-by-line}); and up to ten adaptive rounds raise accuracy for three models but lower it for two, so single-round consultation stays primary. A single frame trails video by $1.4$--$9.9$ points. The clinician's video consultation reaches $35.2\%$ accuracy, second only to GPT-5.6-luna, with fewer questions ($4.6$ versus $6.3$--$16.2$) and investigations ($3.1$ versus $3.5$--$8.5$) than every model. The main contrasts also hold on the $37$ cases published in 2025--2026, where prior exposure is least likely (Appendix~\ref{app:attribution}), on the $49$ clips the neurologist judged identifiable from video alone (Appendix~\ref{app:decidability}), and on cases whose records answer more of the history questions the models asked (Appendix~\ref{app:tau}).

\subsection{Temporal-Window Training and Downstream Diagnosis}
\label{sec:temporal_results}

Post-training improves the sign description, and the better description improves the consultation. Every arm feeds the same GPT-5.6-luna consultation, so only the description differs. A blinded judge prefers the adapter's sentence over the untrained model's for $48$ clips against $14$, mainly because the body part ($40\%$ to $66\%$) and the character of the movement ($10\%$ to $54\%$) are named correctly more often (Appendix~\ref{app:window}). On identical survey-plus-window inputs, the adapter raises source-workup coverage over untrained Qwen3.5-4B by $3.9$--$5.2$ points, with intervals excluding zero at every budget, and accuracy by $5.2$, $8.0$ and $6.6$ points at $K=8,16,32$ (Table~\ref{tab:downstream diagnosis}; Table~\ref{tab:base-vs-adapter}). Sampling densely within a short segment matters more than where the segment lies: Window minus Random accuracy is only $-0.5$, $+1.4$ and $+3.3$ points, yet both windowed arms lead a sign-only adapter that spreads the same $32{+}K$ frame budget over the whole clip by $2.3$--$6.6$ and $1.4$--$3.3$ points at similar coverage (Table~\ref{tab:downstream diagnosis}). A short segment seen at a higher frame rate thus appears more useful than the same frames spread thinly; as with the adapter's accuracy gain, these differences stay within sampling error.

\subsection{Literature Retrieval and Evidence Acquisition}
\label{sec:retrieval_results}

\begin{figure}[H]
\centering
\includegraphics[width=\linewidth]{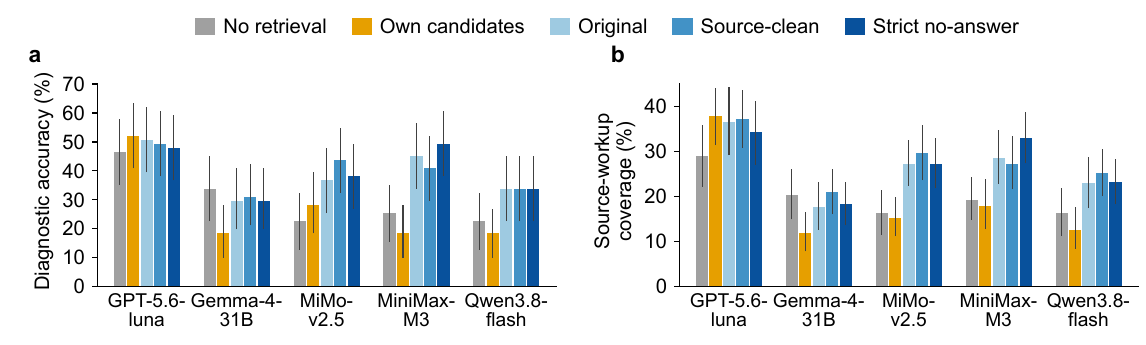}

\caption{Retrieval acts on evidence acquisition.
(a)~Diagnostic accuracy and (b)~source-workup coverage on the 71 cases, with $95\%$ bootstrap intervals.
Own candidates supplies the model's Stage~1 differential alone; retrieval conditions append retrieved causes (Section~\ref{sec:retrieval}): Original keeps all records, Source-clean removes the source article and near duplicates, and Strict no-answer also removes records naming the diagnosis.}
\label{fig:lit}
\end{figure}

Like post-training, retrieval improves diagnosis by acting on evidence acquisition. It first expands the initial hypotheses: the true cause appears in $7.0$--$9.9\%$ of the models' own video-only differentials but in $62.0$--$64.8\%$ of the combined lists supplied to them (Appendix~\ref{app:lit}). With these lists, source-clean retrieval moves the work-up toward the treating clinicians' documented one in four of five systems and significantly raises accuracy for MiMo-v2.5 and MiniMax-M3 (Figures~\ref{fig:retrieval-pipeline} and~\ref{fig:lit}). Much of the accuracy effect comes from the breadth of the list rather than its content: the model's own differential alone (\textbf{Own candidates}) lowers accuracy for three models, appending an equal-length list of unrelated causes (\textbf{Mismatched retrieval}) raises it again, and only MiMo-v2.5 gains from the retrieved content itself, while GPT-5.6-luna gains from neither appended list (Appendix~\ref{app:lit}). Leaked answers do not explain the gains: \textbf{Strict no-answer} retrieval lowers candidate coverage of the true cause by about $13$ points, yet the gains of MiMo-v2.5 and MiniMax-M3 remain, and no retained abstract reports the same patient (Figure~\ref{fig:lit}a; Appendix~\ref{app:lit}). Human re-grading of the source-clean diagnoses reproduces these gains (Appendix~\ref{app:sym-grading}).

\begin{table}[t]
\centering
\caption{Matched LoRA sampling outcomes (Section~\ref{sec:temporal_training}; 71 cases;
three-run per-clip means). Accuracy and source-workup coverage $\tau$ are percentages;
brackets give 95\% clip-level bootstrap intervals (Appendix~\ref{app:window}). No window: sign-only adapter on $32{+}K$ uniform frames (Table~\ref{tab:no-window}). \mbox{W/R/N}: window, random and no-window arms.}
\label{tab:downstream diagnosis}
\small
\setlength{\tabcolsep}{4pt}
\begin{tabular}{@{}ccccc@{}}
\toprule
$K$ & Window accuracy & Random accuracy & No-window accuracy & $\tau$ (W / R / N) \\
\midrule
8 & 29.1 [20.7, 38.0] & 29.6 [21.6, 38.5] & 26.8 [19.2, 34.3] & 17.3 / 16.4 / 19.0 \\
16 & 30.0 [21.1, 39.4] & 28.6 [20.2, 37.6] & 27.2 [19.2, 35.7] & 18.6 / 17.9 / 18.3 \\
32 & 33.8 [24.4, 43.7] & 30.5 [21.6, 39.9] & 27.2 [19.2, 35.7] & 19.4 / 20.9 / 18.9 \\
\bottomrule
\end{tabular}
\par
\parbox{\linewidth}{\small
Baselines (accuracy; coverage). No adapter (K=8/16/32): 23.9 / 22.1 / 27.2; 13.4 / 13.4 / 14.3
(Table~\ref{tab:base-vs-adapter}).}
\end{table}

\section{Conclusion}
\label{sec:conclusion}

\textsc{DynamicDx} evaluates video-based diagnosis as a consultation, scoring each
step from recognizing the sign to ordering tests against a fixed record from the
same case report. Across five vision--language models, video raises accuracy over
blind input, yet recognizing the sign rarely places its cause in the initial
differential and shuffling the frames costs little; the gain arrives instead through
the investigations the video prompts. Models thus appear to read a clip once, as a
description of what it shows, and the decisive step is turning that description into
the right tests: once the decisive investigations are supplied, accuracy reaches
73.2--93.0\%. The two interventions improve what enters
this step: post-training helps a small describer name the sign, especially from a
short, densely sampled segment, and source-clean retrieval widens the initial
differential; both bring the work-up closer to the one clinicians documented and,
through it, improve diagnosis.

For video-based diagnostic models, seeing better helps when it leads to asking
better. Final-answer accuracy hides this route; scoring the work-up against a
fixed record reveals it and compares models on identical evidence. These findings rest on 71 published case reports, whose fixed and partly derived
records cannot reproduce a live consultation and raise absolute accuracy. Extending
the benchmark to other specialties and live recordings would test whether evidence
acquisition remains the bottleneck and whether models that use how a sign unfolds,
not only how it looks, can ease it.

\clearpage
\section*{Disclosure of AI Use}

We used large language models as the systems under evaluation and as tools:
GPT-5.6-luna generated the temporal-window supervision, and DeepSeek-V4.1-Flash
mapped free-text questions and orders to the case records, ran the retrieval
steps and graded outcomes, as described in the methods and appendices.
DeepSeek-V4.1-Flash also generated synthetic data: given
the confirmed diagnosis and the reported chart, it filled the generic panel
of expected investigation values in each chart before
any model was evaluated (Section~\ref{sec:dataset}; Appendix~\ref{app:tau}).
We also used generative AI tools to assist with translation, provide
feedback on methodology, interpret results, check consistency
across the manuscript, revise its structure, wording and LaTeX formatting, and
write plotting code.

The authors reviewed the AI-assisted material and take responsibility
for the final manuscript, including its data, analyses, claims, and
artifacts.

\section*{Ethics Statement}

DynamicDx evaluates diagnostic reasoning using patient presentations
from published case reports. Patient videos may contain identifiable
information; their use and any redistribution require attention to
consent, privacy, and source-specific permissions. The benchmark is
intended for research evaluation, not clinical deployment or autonomous
medical decision-making. Its small, uneven, case-report-derived sample
limits the representativeness of the findings. Human-expert evaluations
and automated grading procedures are described in the paper.

\section*{Reproducibility Statement}

The main text describes benchmark construction, the consultation
protocol, evaluation metrics, and statistical procedures. The
appendices provide case-source attribution, prompts, retrieval and
filtering procedures, temporal-window training settings, and grading
and order-matching audits. Together, these materials document the
experimental setup and support inspection of the reported evaluations. The data, prompts and evaluation code are available at \url{https://github.com/jimmylihui/dynamicDx}.

\bibliography{iclr2025_conference}
\bibliographystyle{iclr2027_conference}

\clearpage
\appendix
\begingroup
\hypersetup{pdfborder={0 0 0}}
\pdfbookmark[0]{Appendix Contents}{appendix.contents}
\begin{center}
    {\large\bfseries Appendix Contents\par}
\end{center}
\vspace{0.5cm}

\newcommand{\appendixcontentsentry}[1]{%
    \hyperref[#1]{\textcolor{red}{\ref*{#1}}} &
    \hyperref[#1]{\textcolor{red}{\nameref*{#1}}} &
    \hyperref[#1]{\textcolor{black}{\pageref*{#1}}}\\[0.35em]}

\noindent
{\fontsize{12}{16}\selectfont\bfseries
\begin{tabularx}{\linewidth}{@{}p{1.5em}>{\raggedright\arraybackslash}Xr@{}}
\appendixcontentsentry{app:attribution}
\appendixcontentsentry{app:decidability}
\appendixcontentsentry{app:tau}
\appendixcontentsentry{app:window}
\appendixcontentsentry{app:frames}
\appendixcontentsentry{app:lit}
\appendixcontentsentry{app:lie}
\appendixcontentsentry{app:prompts}
\appendixcontentsentry{app:temporal_training_details}
\appendixcontentsentry{app:grader}
\end{tabularx}
}
\endgroup
\clearpage

\raggedbottom

\section{Source Attribution and Dataset Characteristics}
\label{app:attribution}

The $71$ cases are drawn from $66$ open-access articles and their
supplementary patient videos. Table~\ref{tab:attribution} lists the source
articles and their licences, retrieved from Europe PMC at build time, together
with the country of the first author's affiliation and the category, age and sex of
each patient filmed. Each PMCID links to its record at
\url{https://europepmc.org/article/PMC/<PMCID>}.

\begin{figure}[H]
    \centering
    \includegraphics[width=\linewidth]{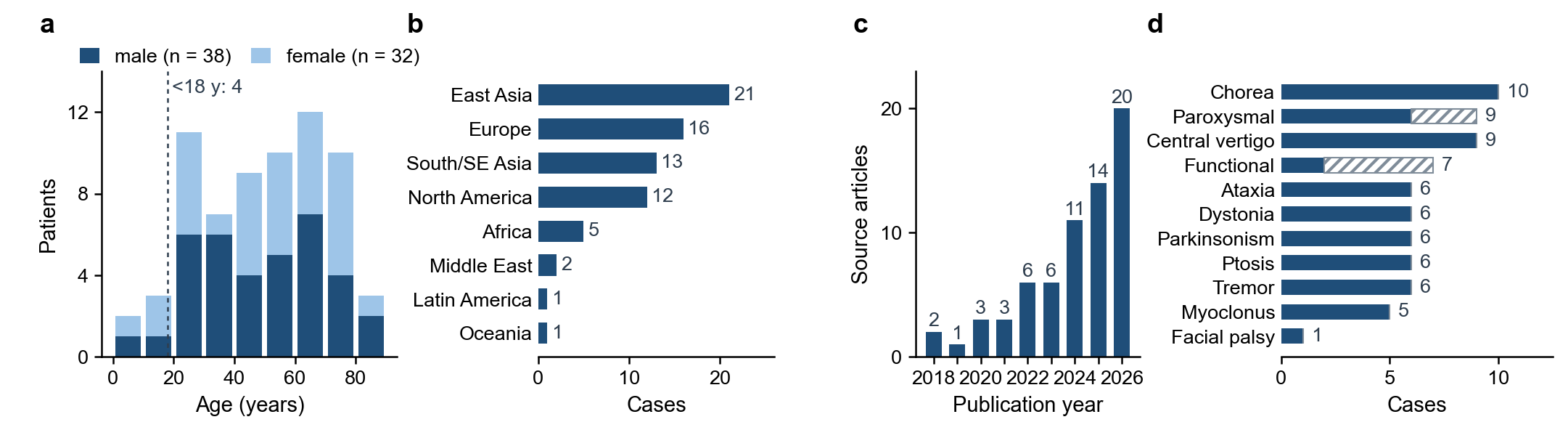}
    \caption{\textbf{Who is in the benchmark.} \textbf{a}, Age of the $67$ patients whose age the
    source states, stacked by sex; four further patients are described only as adults and one has
    no stated sex. The dashed line marks $18$ years. \textbf{b}, Region of the first author's
    affiliation, by case ($23$ countries). \textbf{c}, Publication year of the $66$ source
    articles. \textbf{d}, Cases per category; hatched segments are the eight cases drawn from
    the three articles that contribute more than one patient.}
    \label{fig:dataset-demographics}
\end{figure}

\paragraph{Clustering.} Resampling articles instead of cases leaves the intervals essentially unchanged. Sixty-three articles contribute one case each; three contribute more than one, each clip showing a different patient: a two-patient functional-head-tremor series (\texttt{PMC12999016}), a three-patient functional-gait video series (\texttt{PMC7455329}) and a three-patient ictal-body-turning series (\texttt{PMC13101308}). Intervals that resample the $66$ articles have endpoints within $1$ percentage point of the case-level intervals and nearly the same widths, for every system and the clinician (Table~\ref{tab:cluster-ci}).

\paragraph{Demographics.}
Figure~\ref{fig:dataset-demographics} summarises the sample. Age is stated for $67$ patients (median $52$, IQR $30$--$66$, range $2$--$90$);
four are described only as adults. Four patients are under $18$ (ages $2$, $4$,
$15$ and $17$), $19$ are aged $18$--$39$, $19$ are $40$--$59$, $19$ are $60$--$74$
and $6$ are $75$ or older. Thirty-eight are male, $32$ female and one is not
stated; for three, sex is not given in the text and was read off the video. The
sample is thus mainly adult, with too few patients under $18$ for a subgroup estimate.

\paragraph{Geography and venue.}
By first-author affiliation the $66$ articles come from $23$ countries: China
($13$), India ($11$), the United States ($9$ articles, $12$ cases), Japan ($4$),
South Korea ($4$), Italy ($3$), and $17$ countries with one or two each. By
region the cases are East Asia $21$, Europe $16$, South and South-East Asia $13$,
North America $12$, Africa $5$, Middle East $2$, Latin America $1$ and Oceania
$1$. The articles appear in $36$ journals, most often \emph{Cureus} ($9$),
\emph{Annals of Indian Academy of Neurology} ($5$) and \emph{Journal of Movement
Disorders} ($4$). Publication years (2018--2026) skew recent by
construction of the search: $34$ of $66$ articles are from 2025--2026,
$17$ from 2023--2024 and $15$ earlier.

\paragraph{Diagnostic labels.} The $71$ confirmed diagnoses are almost all distinct entities. The only repeated labels are three ictal-body-turning generalized epilepsies from one series, two functional head tremors from one series, four hyperglycaemia-related choreas, two Graves'-related choreas and two myasthenia gravis presentations; every other case is the sole example of its diagnosis. The case-report sample emphasizes rare and atypical presentations that often require a specific confirming test. The facial-palsy category has one clip, so its statistics carry no interval.

\paragraph{Performance by subgroup.} Publication period and region show the largest gradients in video-condition accuracy (Table~\ref{tab:subgroups}): every model solves articles from 2022 or earlier more often ($29$--$71\%$) than articles from 2023--2024 ($12$--$29\%$), and so does the clinician ($47\%$ vs.\ $18\%$). Pre-training exposure and presentation typicality could both contribute. Cases from South and South-East Asian articles ($n=13$) are also easier for every system and the clinician ($46$--$69\%$, against $16$--$43\%$ for the other $n=58$), but these groups differ in case mix, including Graves'-related and hyperglycaemic chorea and dopa-responsive dystonia. No sex difference is consistent across systems. Because subgroups differ in difficulty, the main analyses compare conditions within the same cases.

\paragraph{Main contrasts by publication period.} Earlier publication makes cases easier once the video is shown, not from the diagnosis alone. For articles from 2022 or earlier, blind accuracy is $16.5\%$, close to $14.6\%$ for 2025--2026, whereas video raises it by $32.9$ points against $10.3$ (Table~\ref{tab:by-period}). The main contrasts hold in the most recent period, where prior exposure is least likely: on the $37$ cases from 2025--2026, video exceeds blind input by $10.3$ points, oracle evidence exceeds video by $54.1$ points and the reference description adds $22.7$ points. Source-clean retrieval adds $14.1$ points in this period and $2.4$ points for articles from 2022 or earlier, consistent with retrieval widening hypotheses most where models know least. Prior exposure may contribute to the advantage of older cases, but it does not produce the main conclusions.

\begin{table}[t]
\centering
\caption{\textbf{Main contrasts by publication period of the source article.} Stage-2 diagnosis accuracy (\%) averaged over the five systems; differences in percentage points with $95\%$ paired case-level bootstrap intervals. Clean: source-clean retrieval.}
\label{tab:by-period}
\small
\setlength{\tabcolsep}{3pt}
\resizebox{\linewidth}{!}{%
\begin{tabular}{lrccccccc}
\toprule
period & $n$ & video & video $-$ blind & oracle $-$ video & reference $-$ video & video $-$ single & video $-$ shuffled & clean $-$ video \\
\midrule
All & 71 & 30.1 & $+16.3$ $[+9.3, +23.7]$ & $+51.8$ $[+43.4, +60.0]$ & $+15.8$ $[+8.7, +22.8]$ & $+5.9$ $[-0.3, +12.4]$ & $+5.4$ $[-0.3, +11.3]$ & $+9.6$ $[+2.5, +16.6]$ \\
$\leq$2022 & 17 & 49.4 & $+32.9$ $[+14.1, +49.4]$ & $+38.8$ $[+23.5, +55.3]$ & $+12.9$ $[-1.2, +27.1]$ & $+16.5$ $[+3.5, +31.8]$ & $+7.1$ $[-5.9, +21.2]$ & $+2.4$ $[-15.3, +21.2]$ \\
2023--24 & 17 & 22.4 & $+12.9$ $[+2.4, +23.6]$ & $+60.0$ $[+43.5, +76.5]$ & $+3.5$ $[-8.2, +16.5]$ & $+3.5$ $[-7.1, +14.1]$ & $+12.9$ $[+5.9, +21.2]$ & $+7.1$ $[-5.9, +21.2]$ \\
2025--26 & 37 & 24.9 & $+10.3$ $[+1.6, +18.9]$ & $+54.1$ $[+42.2, +65.4]$ & $+22.7$ $[+13.5, +32.4]$ & $+2.2$ $[-5.9, +10.8]$ & $+1.1$ $[-7.0, +9.2]$ & $+14.1$ $[+5.4, +22.7]$ \\
\bottomrule
\end{tabular}}
\end{table}

\paragraph{Distribution.}
The benchmark distributes annotations, structured patient and chart tables,
and a script for retrieving the supplementary videos and reproducing the
excerpts locally. It does not redistribute the trimmed and re-encoded
clips: these are derivative works, which the CC~BY-NC-ND licences of
$18$ source articles do not permit us to distribute.

\begin{table}[t]
\centering
\caption{\textbf{Article-clustered bootstrap intervals} for diagnosis accuracy in the video
condition, $10{,}000$ resamples. Clustered resampling draws the $66$ source articles with
replacement, keeping all their cases; the width ratio divides the clustered interval's
width by the case-level one in Table~\ref{tab:consultation_conditions}.}
\label{tab:cluster-ci}
\resizebox{\linewidth}{!}{%
\begin{tabular}{lcccccc}
\toprule
 & GPT-5.6-luna & Gemma-4-31B & MiMo-v2.5 & MiniMax-M3 & Qwen3.8-flash & Clinician \\
\midrule
95\% CI by article & [34.8, 58.1] & [22.9, 45.3] & [13.0, 33.3] & [15.7, 35.7] & [13.6, 32.4] & [24.2, 46.5] \\
width ratio        & 1.04 & 1.00 & 1.03 & 1.01 & 0.95 & 0.99 \\
\bottomrule
\end{tabular}%
}
\end{table}

\begin{table}[t]
\centering
\caption{\textbf{Diagnosis accuracy (\%) in the video condition by subgroup.} The denominator is the
number of cases in the subgroup; a case without a usable answer counts as wrong. Strata with
$n<10$ are shown for completeness only.}
\label{tab:subgroups}
\small
\setlength{\tabcolsep}{4pt}
\begin{tabular}{lrcccccc}
\toprule
subgroup & $n$ & \shortstack{GPT-5.6-\\luna} & \shortstack{Gemma-4-\\31B} & \shortstack{MiMo-\\v2.5} & \shortstack{MiniMax-\\M3} & \shortstack{Qwen3.8-\\flash} & Clinician \\
\midrule
male                 & 38 & 47.4 & 34.2 & 23.7 & 28.9 & 18.4 & 39.5 \\
female               & 32 & 43.8 & 34.4 & 21.9 & 21.9 & 25.0 & 28.1 \\
\midrule
age $<$18            &  4 & 50.0 & 50.0 & 25.0 & 50.0 & 25.0 & 50.0 \\
age 18--59           & 38 & 42.1 & 34.2 & 21.1 & 18.4 & 23.7 & 28.9 \\
age $\geq$60         & 25 & 44.0 & 28.0 & 28.0 & 28.0 & 16.0 & 36.0 \\
\midrule
East Asia            & 21 & 42.9 & 33.3 & 19.0 & 28.6 &  9.5 & 19.0 \\
South/SE Asia        & 13 & 61.5 & 53.8 & 46.2 & 46.2 & 53.8 & 69.2 \\
Europe               & 16 & 50.0 & 37.5 & 18.8 & 18.8 & 31.2 & 43.8 \\
North America        & 12 & 41.7 & 25.0 & 16.7 & 16.7 & 16.7 & 25.0 \\
\shortstack[l]{Africa, Middle East,\\ Latin America, Oceania} & 9 & 33.3 & 11.1 & 11.1 & 11.1 & 0.0 & 22.2 \\
\midrule
published $\leq$2022 & 17 & 70.6 & 52.9 & 47.1 & 47.1 & 29.4 & 47.1 \\
published 2023--24   & 17 & 29.4 & 29.4 & 23.5 & 17.6 & 11.8 & 17.6 \\
published 2025--26   & 37 & 43.2 & 27.0 & 10.8 & 18.9 & 24.3 & 37.8 \\
\midrule
single-case article  & 63 & 42.9 & 34.9 & 25.4 & 23.8 & 22.2 & 31.7 \\
multi-case article   &  8 & 75.0 & 25.0 &  0.0 & 37.5 & 25.0 & 62.5 \\
\bottomrule
\end{tabular}
\end{table}

\begingroup
\small
\setlength{\tabcolsep}{3pt}
\newcolumntype{L}[1]{>{\raggedright\arraybackslash}p{#1}}
\begin{longtable}{L{0.39\linewidth}L{0.14\linewidth}L{0.12\linewidth}L{0.10\linewidth}L{0.17\linewidth}}
\caption{\textbf{Source articles.} Licence, country of the first author's affiliation, and the category, age and sex of each patient filmed (\,--\,: not stated in the source).}\label{tab:attribution}\\
\toprule
Source & PMCID & Licence & Country & Patient(s) \\
\midrule
\endfirsthead
\multicolumn{5}{l}{\small\itshape Table~\ref{tab:attribution}, continued}\\
\toprule
Source & PMCID & Licence & Country & Patient(s) \\
\midrule
\endhead
\bottomrule
\endfoot
Abbassi O et al., Cureus, 2024 & PMC11283634 & CC BY & Morocco & chorea, 66F \\
Antova I et al., Cureus, 2025 & PMC12596229 & CC BY & Bulgaria & chorea, 62M \\
Bartl M et al., Neurological research and practice, 2020 & PMC7713151 & CC BY & Germany & functional, 48F \\
Batot C et al., Tremor and other hyperkinetic movements, 2022 & PMC9122005 & CC BY & France & chorea, 90M \\
Calvo PA et al., Cureus, 2024 & PMC11586875 & CC BY & Portugal & chorea, 84F \\
Chen W et al., Frontiers in neurology, 2022 & PMC9815763 & CC BY & China & chorea, 44F \\
Cutellè R et al., Epileptic disorders : international epilepsy journal with videotape, 2026 & PMC13276695 & CC BY & Italy & paroxysmal, 48M \\
Davis K et al., Journal of education \& teaching in emergency medicine, 2023 & PMC10414977 & CC BY & USA & ptosis, 46F \\
Dentoni M et al., Cerebellum, 2024 & PMC11585521 & CC BY & Italy & ataxia, 71F \\
Doumbouya I et al., Cureus, 2023 & PMC10051035 & CC BY & Guinea & chorea, 64F \\
He Y et al., Frontiers in pediatrics, 2025 & PMC12440885 & CC BY & China & paroxysmal, 4F \\
Horinouchi T et al., Clinical case reports, 2018 & PMC6230673 & CC BY & Japan & paroxysmal, 23M \\
Hu Q et al., Frontiers in genetics, 2025 & PMC12558638 & CC BY & China & tremor, 28M \\
Inzirillo K et al., Cureus, 2025 & PMC12572359 & CC BY & USA & facial palsy, 53M \\
Jimsheleishvili S et al., Neurological sciences : official journal of the Italian Neurological Society and of the Italian Society of Clinical Neurophysiology, 2025 & PMC12678565 & CC BY & USA & ataxia, 59F \\
Liang W et al., Medicine, 2026 & PMC13008232 & CC BY & China & ataxia, 53M \\
Liu C et al., Medicine, 2026 & PMC13166699 & CC BY & China & tremor, 66F \\
Liu Y et al., Frontiers in immunology, 2026 & PMC12979508 & CC BY & China & vertigo, 26F \\
Long X., JPRAS open, 2026 & PMC13049410 & CC BY & China & dystonia, 59F \\
Nambiar SV et al., Clinical parkinsonism \& related disorders, 2026 & PMC13080482 & CC BY & India & parkinsonism, 50M \\
Nie J et al., BMC neurology, 2026 & PMC13085407 & CC BY & China & ataxia, 38M \\
Oliveira D et al., Cureus, 2023 & PMC10656108 & CC BY & Portugal & dystonia, 70M \\
Pham K et al., Cureus, 2020 & PMC7055013 & CC BY & USA & chorea, 76F \\
Rohatgi SJ et al., Cureus, 2024 & PMC11098550 & CC BY & India & chorea, 48M \\
Roy S et al., Tremor and other hyperkinetic movements, 2026 & PMC12904113 & CC BY & India & dystonia, 17M \\
Song DJ et al., Frontiers in immunology, 2025 & PMC12833086 & CC BY & China & vertigo, 15F \\
Srimanan W et al., Cureus, 2026 & PMC13186686 & CC BY & Thailand & vertigo, 34M \\
Stephens T et al., Clinical practice and cases in emergency medicine, 2026 & PMC13135425 & CC BY & USA & vertigo, 53F \\
Stone J et al., Movement disorders : official journal of the Movement Disorder Society, 2026 & PMC13206381 & CC BY & UK & functional, ---- \\
Tai XY et al., Movement disorders : official journal of the Movement Disorder Society, 2025 & PMC12371612 & CC BY & UK & parkinsonism, 54M \\
Tchopev ZN et al., Frontiers in neurology, 2018 & PMC5786569 & CC BY & USA & paroxysmal, 36M \\
Tomic S et al., Frontiers in neurology, 2024 & PMC11021691 & CC BY & Croatia & parkinsonism, 70M \\
Totuk Ö et al., Case reports in psychiatry, 2025 & PMC12297136 & CC BY & Turkey & myoclonus, 73M \\
Vaghi G et al., Cerebellum, 2024 & PMC11102397 & CC BY & Italy & vertigo, 71F \\
Wang RY et al., Journal of medical case reports, 2026 & PMC13224678 & CC BY & China & ataxia, 52F \\
Xie N et al., Tremor and other hyperkinetic movements, 2024 & PMC11639689 & CC BY & China & tremor, 20M \\
Zhang Q et al., Journal of medical case reports, 2019 & PMC6883617 & CC BY & China & vertigo, 62M \\
Chang HJ et al., Journal of clinical neurology, 2023 & PMC9982172 & CC BY-NC & Korea & dystonia, 42M \\
Lee GJ et al., Brain \& NeuroRehabilitation, 2022 & PMC9833462 & CC BY-NC & Korea & myoclonus, 29F \\
Lim SY et al., Journal of movement disorders, 2024 & PMC11082598 & CC BY-NC & Malaysia & parkinsonism, 73F \\
Simma K et al., Oxford medical case reports, 2025 & PMC12741441 & CC BY-NC & Morocco & ataxia, 39M \\
Sugiyama A et al., Case reports in neurology, 2023 & PMC10359687 & CC BY-NC & Japan & tremor, 68M \\
Surisetti BK et al., Journal of movement disorders, 2021 & PMC8490194 & CC BY-NC & India & vertigo, 2M \\
Turgunkhujaev O et al., Journal of movement disorders, 2026 & PMC13175740 & CC BY-NC & Russia & parkinsonism, 58F \\
Youn J et al., Journal of movement disorders, 2023 & PMC10548079 & CC BY-NC & Korea & myoclonus, 60M \\
Zhou LC et al., The Journal of international medical research, 2025 & PMC12745546 & CC BY-NC & China & vertigo, 63M \\
Baizabal-Carvallo JF et al., Movement disorders clinical practice, 2025~\textsuperscript{(2 clips)} & PMC12999016 & CC BY-NC-ND & USA & functional, 61M; functional, 83M \\
Değirmenci MFK et al., BMC ophthalmology, 2024 & PMC11495012 & CC BY-NC-ND & Turkey & ptosis, 40F \\
Hormaza-Jaramillo A et al., Clinical neurophysiology practice, 2026 & PMC12933468 & CC BY-NC-ND & Colombia & myoclonus, 30M \\
Jogi H et al., Annals of Indian Academy of Neurology, 2026 & PMC12962376 & CC BY-NC-ND & India & dystonia, 62F \\
Koutsis G et al., Journal of the peripheral nervous system : JPNS, 2026 & PMC12960836 & CC BY-NC-ND & Greece & ptosis, 45F \\
Kuma A et al., Clinical Case Reports, 2026 & PMC13093780 & CC BY-NC-ND & Ethiopia & tremor, 32F \\
Lee D et al., BMC ophthalmology, 2025 & PMC12606889 & CC BY-NC-ND & Korea & ptosis, 24F \\
Mathuram D et al., Annals of Indian Academy of Neurology, 2025 & PMC12798911 & CC BY-NC-ND & India & paroxysmal, 78F \\
Mohammed S et al., Journal of medical case reports, 2024 & PMC11600610 & CC BY-NC-ND & Ethiopia & tremor, 34M \\
Nagabushana D et al., BMC neurology, 2026~\textsuperscript{(3 clips)} & PMC13101308 & CC BY-NC-ND & USA & paroxysmal, 26F; paroxysmal, 24F; paroxysmal, 29M \\
Nakatake N et al., AACE clinical case reports, 2025 & PMC11973688 & CC BY-NC-ND & Japan & chorea, 73F \\
Nonnekes J et al., Neurology, 2020~\textsuperscript{(3 clips)} & PMC7455329 & CC BY-NC-ND & Netherlands & functional, --M; functional, --F; functional, --M \\
Okadome T et al., Epilepsy \& behavior reports, 2022 & PMC9062418 & CC BY-NC-ND & Japan & paroxysmal, 26M \\
Rodin RE et al., Annals of clinical and translational neurology, 2024 & PMC11093232 & CC BY-NC-ND & USA & chorea, 74F \\
Saibaba J et al., Annals of Indian Academy of Neurology, 2026 & PMC12962431 & CC BY-NC-ND & India & parkinsonism, 52M \\
Verma R et al., Journal of neurosciences in rural practice, 2022 & PMC9357509 & CC BY-NC-ND & India & dystonia, 19F \\
Vinny PW et al., Journal of neurosciences in rural practice, 2021 & PMC8064859 & CC BY-NC-ND & India & ptosis, 75M \\
Yeow D et al., Annals of clinical and translational neurology, 2026 & PMC13071100 & CC BY-NC-ND & Australia & myoclonus, 41M \\
Maramattom BV., Annals of Indian Academy of Neurology, 2021 & PMC8061496 & CC BY-NC-SA & India & vertigo, 64M \\
Pedapati R et al., Annals of Indian Academy of Neurology, 2022 & PMC9350803 & CC BY-NC-SA & India & ptosis, 24M \\
\end{longtable}
\endgroup

\section{Decidability of the Visual Task}
\label{app:decidability}

Low sign recognition could mean that models miss a visible sign, that a clip does not show the sign well enough for any observer, or that the reference description draws on the report rather than the video; the inclusion criterion (Section~\ref{sec:dataset}) was applied with the report in hand. A diagnosis-blind decidability label and an audit of the reference descriptions separate the three. Recognition stays low even on clips a neurologist could read, harder clips lower accuracy, and the contrasts between visual conditions do not depend on them.

\paragraph{A diagnosis-blind decidability label.} The clinician's Stage-1 record supplies an independent judgement of decidability. The clinician viewed each clip before receiving any history, investigation results or diagnosis, and for $22$ of $71$ clips recorded the sign as not identifiable (uncertain, normal or none). We designate these $22$ clips as \emph{clinician-undecidable} and the remaining $49$ as \emph{clinician-decidable}. The undecidable clips are distributed across nine of the eleven categories (paroxysmal $4$; dystonia, functional, ptosis and central vertigo $3$ each; chorea and myoclonus $2$ each; ataxia and parkinsonism $1$ each; tremor and facial palsy $0$). The list is released with the benchmark. We retain all $71$ clips in the primary analysis and use this label for stratified results.

\paragraph{Phenomenology recognition.} Table~\ref{tab:decidability} reports Stage-1 recognition and Stage-2 accuracy by decidability. Recognition does not differ between decidable and undecidable clips in any system (differences of $-8.4$ to $+7.9$ points), and stays low ($10.2$--$30.6\%$) even on clinician-decidable clips.

\paragraph{Diagnostic accuracy.} Accuracy is higher on the $49$ clinician-decidable clips than on the $22$ undecidable clips in every visual condition: $32.2\%$ versus $25.5\%$ with video, $26.9\%$ versus $20.0\%$ with shuffled frames, and $27.3\%$ versus $17.3\%$ with a single frame (means over the five systems), with the same direction in $13$ of $15$ system--condition pairs. The clinician, who also had the history and investigations, scores similarly on the two subsets ($34.7\%$ and $36.4\%$). Part of the low model accuracy may therefore reflect hard-to-read clips.

The contrasts between visual conditions do not depend on these clips (Table~\ref{tab:decidable-conditions}). On the $49$ decidable clips, video exceeds shuffled frames by $5.3$ points and a single frame by $4.9$ points, close to the $5.4$ and $5.9$ points on all $71$ clips; shuffled frames and a single frame differ by $0.4$ points. The small advantage of video over a single frame, and the lack of a reliable ordering effect, hold when undecidable clips are removed.

\paragraph{Visual grounding of the reference descriptions.} The reference condition supplies a clinician-written description of the visible sign, prepared with access to the source report. We decomposed all $71$ descriptions into atomic claims and classified each as observable or not observable in a silent video. Of $286$ claims, $34$ ($11.9\%$), in $27$ descriptions, were classified as not observable. Most are temporal or volitional qualifiers such as continuous, involuntary, or sudden. Nine phrases in eight descriptions exceed the visual evidence: reported symptoms (double vision, painful), contextual information (arising from sleep with vocalisation, bed-bound, alert and able to speak), the label seizure in two descriptions, and one editorial remark about lumbar hyperlordosis. We removed these phrases from the released descriptions and updated the reference-condition results and student training targets accordingly. The complete audit is released with the descriptions.

\begin{table}[t]
\centering
\caption{\textbf{Results by clinician-judged decidability.} The neurologist judged $22$ clips
unidentifiable from the video alone, before seeing case material. Recognition is the Stage-1
sign grade at $32$ frames (\emph{correct}); accuracy is Stage-2 video-condition
accuracy. Cells give decidable ($n{=}49$) / undecidable ($n{=}22$); the gap is their
difference with a $10{,}000$-resample bootstrap interval.}
\label{tab:decidability}
\small
\setlength{\tabcolsep}{5pt}
\begin{tabular}{lcc@{\hskip 14pt}cc}
\toprule
 & \multicolumn{2}{c}{sign recognition (\%)} & \multicolumn{2}{c}{diagnosis accuracy (\%)} \\
\cmidrule(lr){2-3}\cmidrule(lr){4-5}
 & dec.\ / undec. & gap [95\% CI] & dec.\ / undec. & gap [95\% CI] \\
\midrule
GPT-5.6-luna  & 30.6 / 22.7 & $+7.9$ $[-14.4, +29.2]$ & 46.9 / 45.5 & $+1.5$ $[-23.7, +26.9]$ \\
Gemma-4-31B   & 14.3 / 22.7 & $-8.4$ $[-28.7, +11.3]$  & 36.7 / 27.3 & $+9.5$ $[-13.7, +31.7]$ \\
MiMo-v2.5     & 26.5 / 22.7 & $+3.8$ $[-18.5, +24.0]$  & 24.5 / 18.2 & $+6.3$ $[-14.8, +26.1]$ \\
MiniMax-M3    & 10.2 / 18.2 & $-8.0$ $[-27.7, +9.7]$  & 30.6 / 13.6 & $+17.0$ $[-2.8, +35.8]$ \\
Qwen3.8-flash & 26.5 / 27.3 & $-0.7$ $[-23.5, +21.1]$  & 22.4 / 22.7 & $-0.3$ $[-22.1, +19.9]$ \\

\bottomrule
\end{tabular}
\end{table}

\begin{table}[t]
\centering
\caption{\textbf{Visual conditions on the $49$ clinician-decidable clips.} Stage-2 diagnosis accuracy (\%) with $32$ ordered frames (video), $32$ shuffled frames, and a single frame, and paired differences with $95\%$ case-level bootstrap intervals. The last row averages the five systems, resampling cases jointly.}
\label{tab:decidable-conditions}
\small
\setlength{\tabcolsep}{4pt}
\resizebox{\linewidth}{!}{%
\begin{tabular}{lccc@{\hskip 10pt}ccc}
\toprule
 & video & shuffled & single & video $-$ shuffled & video $-$ single & shuffled $-$ single \\
\midrule
GPT-5.6-luna  & 46.9 & 51.0 & 44.9 & $-4.1$ $[-16.3, +6.1]$ & $+2.0$ $[-12.2, +16.3]$ & $+6.1$ $[-8.2, +20.4]$ \\
Gemma-4-31B   & 36.7 & 28.6 & 28.6 & $+8.2$ $[-4.1, +20.4]$ & $+8.2$ $[-6.1, +22.4]$ & $+0.0$ $[-14.3, +14.3]$ \\
MiMo-v2.5     & 24.5 & 26.5 & 22.4 & $-2.0$ $[-18.4, +16.3]$ & $+2.0$ $[-14.3, +18.4]$ & $+4.1$ $[-10.2, +18.4]$ \\
MiniMax-M3    & 30.6 & 14.3 & 18.4 & $+16.3$ $[+2.0, +30.6]$ & $+12.2$ $[-2.0, +28.6]$ & $-4.1$ $[-16.3, +8.2]$ \\
Qwen3.8-flash & 22.4 & 14.3 & 22.4 & $+8.2$ $[-2.0, +20.4]$ & $+0.0$ $[-14.3, +14.3]$ & $-8.2$ $[-22.4, +6.1]$ \\
\midrule
Mean of five  & 32.2 & 26.9 & 27.3 & $+5.3$ $[-1.6, +12.2]$ & $+4.9$ $[-3.3, +13.5]$ & $-0.4$ $[-7.8, +6.9]$ \\
\bottomrule
\end{tabular}%
}
\end{table}

\section{Evidence Acquisition and Investigation Selection}
\label{app:tau}

This appendix supports the claim that video helps mainly through the investigation results it obtains (Section~\ref{sec:consultation_results}). Section~\ref{app:evidence-paths} traces the video gain, Section~\ref{app:records} describes what the records contain and how $\tau$ is scored, and Section~\ref{app:selection} compares model-selected work-ups with fixed and random lists.

\subsection{How the Video Gain Reaches the Diagnosis}
\label{app:evidence-paths}

\paragraph{Where the video gain arises.} The video gain is concentrated in consultations where video changes which decisive evidence is obtained. For each system we split the $71$ cases by whether the video consultation obtains more, the same, or fewer decisive entries than the blind consultation, and attribute the accuracy change to each group (Table~\ref{tab:video-decomp}). Pooled over systems, $15.8$ of the $16.3$-point gain comes from the $35\%$ of consultations in which video obtains a decisive entry that blind input missed; within this group accuracy rises by $44.4$ points. Order counts do not rise with video except for Qwen3.8-flash (Table~\ref{tab:availability}), yet consultations with at least one decisive entry rise from $115$ to $188$ of $355$. Consultations that recover more documented history but the same decisive evidence contribute $0.6$ points, whereas those with more decisive evidence contribute $16.0$ points whatever the change in history. The replay below tests these paths directly.

\begin{table}[t]
\centering
\caption{\textbf{Where the video gain arises.} For each system the $71$ cases are split by whether the video consultation obtains more, the same, or fewer distinct decisive entries than the blind consultation. Each cell gives the number of cases and that group's contribution to the video $-$ blind accuracy gain (percentage points of the $71$-case rate); contributions sum to the gain. The last column counts consultations with at least one decisive entry, blind $\rightarrow$ video.}
\label{tab:video-decomp}
\small
\setlength{\tabcolsep}{4pt}
\begin{tabular}{lccccc}
\toprule
system & gain & more decisive & same decisive & fewer decisive & cases with a decisive entry \\
\midrule
GPT-5.6-luna & $+22.5$ & 24: $+16.9$ & 39: $+9.9$ & 8: $-4.2$ & 34 $\rightarrow$ 45 \\
Gemma-4-31B & $+9.9$ & 18: $+11.3$ & 40: $+4.2$ & 13: $-5.6$ & 34 $\rightarrow$ 38 \\
MiMo-v2.5 & $+14.1$ & 27: $+14.1$ & 33: $+2.8$ & 11: $-2.8$ & 16 $\rightarrow$ 33 \\
MiniMax-M3 & $+15.5$ & 29: $+16.9$ & 29: $+2.8$ & 13: $-4.2$ & 26 $\rightarrow$ 42 \\
Qwen3.8-flash & $+19.7$ & 28: $+19.7$ & 40: $+1.4$ & 3: $-1.4$ & 5 $\rightarrow$ 30 \\
\midrule
Pooled & $+16.3$ & 126: $+15.8$ & 181: $+4.2$ & 48: $-3.7$ & 115 $\rightarrow$ 188 \\
\bottomrule
\end{tabular}
\end{table}

\paragraph{Trajectory replay.} At the diagnosis step, the video gain runs mainly through the investigation results the video consultation obtained. We re-ran only the final diagnosis turn, always with the $32$ video frames, while crossing the source of the history (questions and answers) with the source of the investigations (orders and returned results): each came either from the video or from the blind consultation of the same case (Table~\ref{tab:replay}). Replaying the video trajectory reproduces the original video accuracy ($29.3\%$ versus $30.1\%$, means over systems). Replacing the investigations of the video consultation with the blind ones lowers accuracy by $12.1$ points whether the history comes from the video or the blind consultation, whereas replacing the history changes accuracy by $0.3$ points with either set of investigations. Showing the frames at diagnosis to an otherwise blind trajectory adds $3.1$ points. Most of the $15.5$-point replayed gain therefore enters the diagnosis through the returned investigations rather than through the history or a second look at the video.

\begin{table}[t]
\centering
\caption{\textbf{Trajectory replay of the diagnosis turn.} (a)~Accuracy (\%). The original runs are the blind and video consultations; each replayed diagnosis turn sees the $32$ video frames and combines the history of one consultation with the investigations and returned results of another. (b)~Paired differences averaged over the five systems, with $95\%$ case-level bootstrap intervals.}
\label{tab:replay}
\small
\setlength{\tabcolsep}{5pt}
\textbf{(a)}\\[2pt]
\begin{tabular}{lcc@{\hspace{12pt}}cccc}
\toprule
 & \multicolumn{2}{c}{original run} & \multicolumn{4}{c}{replayed diagnosis turn} \\
\cmidrule(lr){2-3}\cmidrule(lr){4-7}
history from & & & video & video & blind & blind \\
investigations from & & & video & blind & video & blind \\
system & blind & video & VV & VB & BV & BB \\
\midrule
GPT-5.6-luna & 23.9 & 46.5 & 42.3 & 31.0 & 39.4 & 32.4 \\
Gemma-4-31B & 23.9 & 33.8 & 35.2 & 28.2 & 31.0 & 28.2 \\
MiMo-v2.5 & 8.5 & 22.5 & 18.3 & 8.5 & 26.8 & 8.5 \\
MiniMax-M3 & 9.9 & 25.4 & 28.2 & 12.7 & 26.8 & 11.3 \\
Qwen3.8-flash & 2.8 & 22.5 & 22.5 & 5.6 & 21.1 & 4.2 \\
\midrule
\textbf{Mean} & \textbf{13.8} & \textbf{30.1} & \textbf{29.3} & \textbf{17.2} & \textbf{29.0} & \textbf{16.9} \\
\bottomrule
\end{tabular}

\vspace{8pt}
\textbf{(b)}\\[2pt]
\begin{tabular}{lll}
\toprule
path & contrast & $\Delta$ accuracy (pp) \\
\midrule
Investigations & VV $-$ VB \ (video history) & $+12.1$ $[+6.5, +17.7]$ \\
 & BV $-$ BB \ (blind history) & $+12.1$ $[+5.9, +18.3]$ \\
History & VV $-$ BV \ (video investigations) & $+0.3$ $[-2.8, +3.1]$ \\
 & VB $-$ BB \ (blind investigations) & $+0.3$ $[-1.7, +2.5]$ \\
Frames at diagnosis & BB $-$ blind & $+3.1$ $[+0.3, +6.5]$ \\
Replay fidelity & VV $-$ video & $-0.8$ $[-4.2, +2.5]$ \\
\bottomrule
\end{tabular}
\end{table}

\paragraph{Evidence-return patterns in incorrect consultations.} Table~\ref{tab:failure-decomp} partitions incorrect video-condition consultations by the investigation results returned: (i) at least one decisive entry; (ii) other results but no decisive entry; and (iii) no returned result. Group (i) accounts for $29$--$49\%$ of incorrect consultations, group (ii) for $34$--$67\%$ and group (iii) for $4$--$29\%$. Missing decisive evidence is therefore common, yet in roughly a third to a half of incorrect consultations at least one decisive entry was returned without leading to the diagnosis.

\begin{table}[H]
\centering
\caption{\textbf{Evidence returned in incorrect video-condition consultations.}
Consultations not graded \textsc{Accurate}, split by the recorded investigation results into
three exclusive categories.}
\label{tab:failure-decomp}
\small
\setlength{\tabcolsep}{5pt}
\begin{tabular}{lcccc}
\toprule
 & incorrect & \multicolumn{3}{c}{of which (\%)} \\
\cmidrule(lr){3-5}
system & (of 71) & decisive entry & no decisive entry, & no results \\
 & & returned & some results returned & returned \\
\midrule
GPT-5.6-luna  & 38 & 37 & 34 & 29 \\
Gemma-4-31B   & 47 & 34 & 60 &  6 \\
MiMo-v2.5     & 55 & 35 & 45 & 20 \\
MiniMax-M3    & 53 & 49 & 45 &  6 \\
Qwen3.8-flash & 55 & 29 & 67 &  4 \\
\bottomrule
\end{tabular}
\end{table}

\subsection{Evidence Available in the Records and the Scoring of \texorpdfstring{$\tau$}{tau}}
\label{app:records}

\paragraph{Chart composition.} Most chart entries are derived rather than reported. Each case chart combines the results its source article reports with the investigation menu of its category. Every case in a category offers the same menu, so that a reasonable test the report omits still returns a result, and the presence of a result does not reveal which tests the report performed; entries a case does not report carry a value expected for its presentation (\texttt{derived}), in most cases normal, not performed or not recorded ($82\%$ of the $5{,}972$ derived entries). Derived values were fixed before any model was evaluated: category-specific entries were written by the annotators, and a generic panel of commonly ordered tests was filled per case with DeepSeek-V4.1-Flash given the confirmed diagnosis and the reported chart, under the instruction to answer normal unless the diagnosis specifically changes the test. The $45$ derived decisive entries are menu tests whose expected result bears on the diagnosis; the annotators flagged them together with the reported ones ($44$ from the category menus, one from the generic panel). Table~\ref{tab:chart-composition} gives the counts.

\begin{table}[H]
\centering
\caption{\textbf{Chart composition} over the $71$ cases. Neutral: normal, negative, not performed, not recorded or not tried; specific: any other value.}
\label{tab:chart-composition}
\small
\begin{tabular}{lrrr}
\toprule
source of the entry & entries & decisive & specific values \\
\midrule
reported in the source article & $426$ & $256$ & $394$ \\
derived, category menu (annotators) & $2{,}473$ & $44$ & $808$ \\
derived, generic panel (DeepSeek-V4.1-Flash) & $3{,}499$ & $1$ & $278$ \\
\midrule
total & $6{,}398$ & $301$ & $1{,}480$ \\
\bottomrule
\end{tabular}
\end{table}

\paragraph{Sensitivity to derived values.} Most of the video benefit, and all of the oracle and retrieval effects, survive the removal of derived values. We replaced each of the $1{,}086$ derived entries that hold a specific value with the neutral value a sibling case returns (normal / non-contributory, not performed, not recorded or not tried) and re-ran only the diagnosis turn, holding each consultation's questions, orders and matched entries fixed; because every order is placed before any result is returned, this equals a full re-run of the consultation. Neutral values lower accuracy by about ten points with video and five without, so derived values raise absolute accuracy. Video stays ahead of blind input by $10.1$ points ($14.9$ with the original chart), the oracle reference is unchanged, and retrieval keeps its advantage over video (Table~\ref{tab:neutral}). Derived values therefore carry about a third of the video benefit and none of the oracle or retrieval effects.

\begin{table}[H]
\centering
\caption{\textbf{Neutral-value sensitivity.} Accuracy (\%), mean of five systems, with the diagnosis turn re-run on fixed trajectories under the original chart and with every derived specific value neutralised. Differences are pooled over systems with $95\%$ paired case-level bootstrap intervals; all rows use the $71$ cases, with an empty reply counted as incorrect.}
\label{tab:neutral}
\small
\begin{tabular}{lrrr}
\toprule
 & original & neutral & neutral $-$ original \\
\midrule
Blind & $14.4$ & $9.3$ & $-5.1$ $[-8.2, -2.3]$ \\
Video & $29.3$ & $19.4$ & $-9.9$ $[-13.5, -6.2]$ \\
Oracle evidence & $81.4$ & $80.0$ & $-1.4$ $[-5.4, +2.5]$ \\
Source-clean retrieval & $39.7$ & $30.4$ & $-9.3$ $[-13.5, -5.1]$ \\
\midrule
Video $-$ blind & $+14.9$ & $+10.1$ & $-4.8$ $[-9.3, -0.3]$ \\
Retrieval $-$ video & $+10.4$ & $+11.0$ & $+0.6$ $[-4.8, +5.9]$ \\
\bottomrule
\end{tabular}
\end{table}

\paragraph{Availability of requested evidence.}
The records settle few of the questions asked (Table~\ref{tab:availability}). Depending on system and condition, $67$--$92\%$ of questions are answered
\emph{unknown} and $7$--$52\%$ of orders return nothing. The reference condition
has lower unavailability than the video condition for all five systems.
Compared with the video condition, the multi-turn protocol reduces order
unavailability for GPT-5.6-luna, MiMo-v2.5, and Qwen3.8-flash, but increases
it for Gemma-4-31B and MiniMax-M3. These rates depend on both the model's
requests and the record: the same $71$ records settle between $8\%$ and
$33\%$ of the questions and answer between $48\%$ and $93\%$ of the orders.

\begin{table}[tbp]
\centering
\caption{\textbf{Availability of requested evidence by condition.} Per-case means over the usable consultations of the $71$ clips; $\tau$ as in Table~\ref{tab:consultation_conditions}. History reports questions and unknown answers; investigations report orders, unavailable returns, distinct entries released, reported or derived (Table~\ref{tab:chart-composition}), and $\tau$. Multi-turn uses up to ten rounds; reference uses audited descriptions.}
\label{tab:availability}
\footnotesize
\setlength{\tabcolsep}{4pt}
\textit{History} (questions asked per case; share answered ``unknown'', \%)\par\smallskip
\begin{tabular}{ll@{\hskip 8pt}rrrrrr}
\toprule
system & & blind & single frame & shuffled & video & multi-turn & reference \\
\midrule
\multirow{2}{*}{GPT-5.6-luna} & questions & 29.8 & 14.9 & 17.6 & 16.2 & 25.1 & 18.5 \\
& unknown & 92.1 & 86.5 & 84.8 & 85.7 & 84.8 & 83.3 \\
\addlinespace
\multirow{2}{*}{Gemma-4-31B} & questions & 4.7 & 9.0 & 8.0 & 7.9 & 23.5 & 6.8 \\
& unknown & 82.5 & 81.1 & 77.7 & 77.5 & 84.4 & 66.6 \\
\addlinespace
\multirow{2}{*}{MiMo-v2.5} & questions & 12.4 & 8.2 & 7.1 & 8.2 & 19.8 & 9.3 \\
& unknown & 89.3 & 88.7 & 84.2 & 84.2 & 86.4 & 70.5 \\
\addlinespace
\multirow{2}{*}{MiniMax-M3} & questions & 19.6 & 15.9 & 11.6 & 11.3 & 26.3 & 12.1 \\
& unknown & 88.7 & 88.9 & 86.1 & 84.6 & 86.1 & 73.8 \\
\addlinespace
\multirow{2}{*}{Qwen3.8-flash} & questions & 7.9 & 4.0 & 6.0 & 6.3 & 8.4 & 7.0 \\
& unknown & 91.9 & 88.3 & 86.0 & 87.9 & 88.4 & 77.8 \\
\bottomrule
\end{tabular}
\par\medskip
\textit{Investigations} (orders per case; share unavailable, \%; entries released per case; $\tau$, \%)\par\smallskip
\begin{tabular}{ll@{\hskip 8pt}rrrrrr}
\toprule
system & & blind & single frame & shuffled & video & multi-turn & reference \\
\midrule
\multirow{4}{*}{GPT-5.6-luna} & orders & 8.7 & 7.9 & 9.2 & 8.5 & 6.8 & 8.0 \\
& unavailable & 42.8 & 29.0 & 19.1 & 35.0 & 20.8 & 14.8 \\
& entries & 15.4 & 16.9 & 21.2 & 16.7 & 13.0 & 22.3 \\
& $\tau$ & 19.3 & 28.6 & 32.9 & 28.9 & 39.2 & 50.8 \\
\addlinespace
\multirow{4}{*}{Gemma-4-31B} & orders & 3.9 & 4.1 & 3.7 & 3.5 & 4.9 & 2.6 \\
& unavailable & 13.6 & 23.3 & 16.2 & 17.8 & 27.1 & 6.7 \\
& entries & 17.6 & 10.5 & 8.0 & 8.2 & 6.2 & 6.9 \\
& $\tau$ & 22.3 & 17.3 & 15.0 & 20.3 & 17.6 & 19.9 \\
\addlinespace
\multirow{4}{*}{MiMo-v2.5} & orders & 6.8 & 6.5 & 5.5 & 6.2 & 5.1 & 5.6 \\
& unavailable & 52.2 & 31.6 & 23.8 & 25.9 & 22.0 & 15.3 \\
& entries & 8.9 & 9.0 & 8.2 & 9.0 & 5.5 & 9.2 \\
& $\tau$ & 8.3 & 16.9 & 18.9 & 16.3 & 14.0 & 22.3 \\
\addlinespace
\multirow{4}{*}{MiniMax-M3} & orders & 9.2 & 12.6 & 9.7 & 7.7 & 7.2 & 9.0 \\
& unavailable & 36.7 & 42.2 & 31.0 & 27.3 & 32.1 & 22.8 \\
& entries & 12.0 & 8.3 & 9.7 & 8.4 & 7.3 & 10.7 \\
& $\tau$ & 13.6 & 11.0 & 15.6 & 19.3 & 21.6 & 30.6 \\
\addlinespace
\multirow{4}{*}{Qwen3.8-flash} & orders & 3.5 & 3.7 & 4.8 & 5.3 & 2.4 & 2.8 \\
& unavailable & 37.6 & 34.5 & 23.4 & 21.9 & 11.8 & 13.3 \\
& entries & 4.0 & 5.9 & 6.1 & 7.9 & 4.8 & 4.5 \\
& $\tau$ & 2.0 & 16.3 & 13.6 & 16.3 & 10.6 & 17.9 \\
\bottomrule
\end{tabular}
\end{table}

\paragraph{Answerability of each record.} Record sparsity does not drive the main contrasts. We define a case's answerability as the share of all questions asked about it, pooled over the five systems and six conditions of Table~\ref{tab:availability}, that its record settles. Answerability ranges from $5.7\%$ to $28.2\%$ across cases (median $14.7\%$) and is unrelated to the number of documented history features ($r=-0.02$), so sparsity reflects a mismatch between the questions asked and what reports record rather than short records. On the $35$ cases with above-median answerability, video exceeds blind input by $21.7$ points, oracle evidence exceeds video by $48.6$ points and the reference description adds $13.1$ points, against $16.3$, $51.8$ and $15.8$ on all $71$ cases; video and single-frame accuracy differ by $3.4$ points (Table~\ref{tab:answerability}). Video-condition accuracy is the same in both halves ($30.3\%$ and $30.0\%$). The conclusions therefore hold where the record answers more of what is asked.

\begin{table}[t]
\centering
\caption{\textbf{Main contrasts by record answerability.} Answerability is the share of a case's questions, pooled over five systems and six conditions, that its record settles (median $14.7\%$). Stage-2 accuracy (\%) averaged over systems; differences in points with $95\%$ paired case-level bootstrap intervals.}
\label{tab:answerability}
\small
\setlength{\tabcolsep}{3pt}
\resizebox{\linewidth}{!}{%
\begin{tabular}{lrcccccc}
\toprule
cases & $n$ & video & video $-$ blind & oracle $-$ video & reference $-$ video & video $-$ single & video $-$ shuffled \\
\midrule
All cases & 71 & 30.1 & $+16.3$ $[+9.3, +23.7]$ & $+51.8$ $[+43.4, +60.0]$ & $+15.8$ $[+8.7, +22.8]$ & $+5.9$ $[-0.3, +12.4]$ & $+5.4$ $[-0.3, +11.3]$ \\
Above-median answerability & 35 & 30.3 & $+21.7$ $[+13.1, +30.9]$ & $+48.6$ $[+36.6, +61.1]$ & $+13.1$ $[+2.9, +23.4]$ & $+3.4$ $[-5.1, +12.0]$ & $+3.4$ $[-5.1, +12.0]$ \\
At or below median & 36 & 30.0 & $+11.1$ $[+0.0, +22.2]$ & $+55.0$ $[+43.9, +66.1]$ & $+18.3$ $[+8.9, +27.8]$ & $+8.3$ $[+0.0, +17.2]$ & $+7.2$ $[-0.6, +14.4]$ \\
Top third & 24 & 31.7 & $+22.5$ $[+11.7, +34.2]$ & $+49.2$ $[+33.3, +65.0]$ & $+15.8$ $[+3.3, +28.3]$ & $+0.0$ $[-10.0, +10.0]$ & $+3.3$ $[-7.5, +15.0]$ \\
\bottomrule
\end{tabular}}
\end{table}

\paragraph{Acquisition of documented history.} In the video condition, models acquire only $15$--$35\%$ of the positive history findings the records document (Table~\ref{tab:history-acq}). Every case documents $15$--$24$ history features (median $18$), of which $5$--$13$ are recorded as present (median $8$); a positive finding counts as acquired when a model question would reveal it, with a judge mapping questions to the symptom table. Video raises acquisition over blind input for every system, and multi-turn interaction raises it again; paired comparisons support both gains. The video gain appears in all $15$ system--stratum cells when cases are split by the number of documented positive findings ($\le 7$, $8$, $\ge 9$; $n = 29, 17, 25$). Across the $1{,}065$ consultations, accuracy is $20.1\%$ when fewer than a quarter of the positive findings are acquired, $33.9\%$ at $25$--$50\%$, and $30.6\%$ above half.

\begin{table}[t]
\centering
\caption{\textbf{Acquisition of documented history}: share (\%) of the findings each record
documents as present that the consultation's questions would have revealed, pooled over $71$
cases, with $95\%$ case-level bootstrap intervals; differences are paired over cases.}
\label{tab:history-acq}
\small
\setlength{\tabcolsep}{3pt}
\begin{tabular*}{\linewidth}{@{\extracolsep{\fill}}lccccc@{}}
\toprule
system & blind & video & multi-turn & video $-$ blind & multi-turn $-$ video \\
\midrule
GPT-5.6-luna  & 19.0 & 34.8 & 43.6 & $+15.9$ $[+8.9, +23.2]$ & $+8.8$ $[+1.4, +16.3]$ \\
Gemma-4-31B   &  7.9 & 19.5 & 31.6 & $+11.6$ $[+6.0, +17.4]$ & $+12.1$ $[+4.7, +19.3]$ \\
MiMo-v2.5     & 14.7 & 23.8 & 38.8 & $+9.1$ $[+1.6, +16.7]$  & $+15.0$ $[+6.3, +23.8]$ \\
MiniMax-M3    & 11.9 & 21.2 & 37.6 & $+9.3$ $[+4.2, +14.7]$  & $+16.4$ $[+9.4, +23.2]$ \\
Qwen3.8-flash &  7.4 & 14.8 & 20.9 & $+7.4$ $[+2.5, +12.4]$  & $+6.0$ $[+0.3, +11.8]$ \\
\bottomrule
\end{tabular*}
\end{table}

\paragraph{Scoring of $\tau$.} For case $c$, with decisive entries $T_c$ and the chart entries $E_{\mathrm{acq},c}$ its consultation acquires, deduplicated within the case, $\tau_c=|T_c\cap E_{\mathrm{acq},c}|/|T_c|$. Entries are equally weighted and may support the diagnosis or exclude alternatives. Reported values pool over the $71$ cases, $\tau_{\mathrm{pooled}}=\sum_c |T_c\cap E_{\mathrm{acq},c}|/\sum_c |T_c|$. A decisive entry counts as acquired when it, or another entry of the same chart that reports the same finding from the same kind of investigation, is returned; these equivalences are fixed per case before scoring, as described next. $\tau$ measures how much of the documented work-up a consultation reproduces; blinded ratings of the order lists are in Appendix~\ref{app:selection}.

\paragraph{Equivalent chart entries.} Scoring decisive entries by name alone undercounts evidence, so $\tau$ credits equivalent entries. Under name-only scoring, $22$ of the $107$ accurate video-condition consultations obtained no decisive entry, and a verifier review found that in $15$ of them the decisive finding had been returned under a second entry for the same test (Table~\ref{tab:tau0-audit}); for example, the hummingbird sign came back under ``MRI brain with gadolinium'' while the decisive entry is ``MRI brain -- midbrain''. Before scoring, we therefore listed for every case the chart entries that report a decisive finding from the same kind of investigation. A judge proposed links from each full chart, and a second, pairwise pass kept $90$ of $255$ proposals, rejecting normal, less specific, merely compatible and cross-modality results. The links recover $12$ of the $15$ audited duplicates and none of the other audited consultations. Under this scoring $10$ accurate consultations retain $\tau=0$, $\tau$ rises by $0.4$--$7.0$ points across the conditions of Table~\ref{tab:consultation_conditions}, and the main-text paired contrasts keep their direction.

\begin{table}[h]
\centering
\caption{\textbf{Accurate video-condition consultations with name-only $\tau=0$} ($22$ of $107$), classified by a verifier review of the orders, returned results and source report.}
\label{tab:tau0-audit}
\small
\begin{tabular}{p{0.46\linewidth}cp{0.36\linewidth}}
\toprule
category & $n$ & example \\
\midrule
Decisive finding returned under a second entry for the same test & 15 & PSP: hummingbird sign under ``MRI brain with gadolinium'' \\
Different but reasonable diagnostic route & 3 & mucormycosis: sinonasal--orbital disease on MRI rather than CT \\
Decisive order lost to a matching miss & 1 & ``MRI Brain'' not matched to the pontine-infarct MRI \\
No supporting result; diagnosis from video and history & 3 & post-hypoxic myoclonus without the cardiac-arrest history \\
\bottomrule
\end{tabular}
\end{table}

\FloatBarrier
\subsection{Case-Specific Investigation Selection}
\label{app:selection}

Model-selected work-ups beat a random list of the same length but not a fixed ten-item checklist, whose accuracy and decisive-evidence coverage are close to model selection. All arms use the same case records; accuracy, $\tau$ and blinded list ratings capture different properties of the work-up.

\paragraph{Experimental controls.}
We replay all $71$ clips on GPT-5.6-luna, keeping the video, history questions
and patient responses fixed. Only the investigation turn changes, followed
by diagnosis using the returned evidence. The \emph{budget} arm
allows ten atomic tests, with one test per line and no bundled or generic
requests such as ``routine bloods''. The \emph{checklist} arms submit the
same fixed list of ten or fifteen items for every case. A \emph{random} arm
submits ten entry names sampled uniformly from the pooled chart vocabulary.
Atomic wording fixes the request, not the release: one free-text request can
still match several chart entries (the released video arms return $1.1$ to
$2.4$ entries per order, and the budget arm's ten atomic requests release
$1.53$ entries per answered order). Two \emph{atomic checklist} arms therefore
hold the ten-item checklist to one chart entry per order. In the \emph{named} variant, each item
is resolved to a single named chart entry of the case and only that entry is
released; an item with no such entry returns ``not performed / not
available''. In the \emph{matched} variant, the checklist is submitted as
written and the matcher releases at most one entry per order, the entry that
most exactly names the test. A third single-entry arm resubmits the
budget arm's own ten orders under the same rule, so both sides
use one release rule. All arms use the same matcher and case-specific
chart.

The ten-item checklist comprises tone; power; eye movements; full blood
count; electrolytes and renal function; liver function; glucose and HbA1c;
thyroid function; vitamin B12; and contrast brain MRI. The fifteen-item list
adds gait; copper and caeruloplasmin; EEG; nerve conduction studies with
electromyography; and lumbar puncture.

\begin{table}[htbp]
\centering
\caption{\textbf{Order-selection controls} on GPT-5.6-luna across $71$ clips.
Orders denotes mean orders per clip; unav.\ is the share of orders returning
nothing. Entries denotes mean distinct chart entries released per clip,
deduplicated across orders within each case (entries divided by orders gives
the entries released per order). Brackets show $95\%$ case-level bootstrap intervals. $\Delta$ is the paired
difference in accuracy relative to the released free-text arm, which is the
video arm of Table~\ref{tab:consultation_conditions}. $\tau$ as in Section~\ref{sec:measurements}. The two atomic
checklists hold the ten-item checklist to one chart entry per order:
\emph{named} resolves each item to a single named chart entry of the case,
\emph{matched} submits the checklist as written and lets the matcher release
at most one entry per order. \emph{Budget, single entry} releases the
budget arm's own orders under the matched rule. Blinded ratings of these lists are in
Table~\ref{tab:app-rating}.}
\label{tab:app-selection}
\footnotesize
\setlength{\tabcolsep}{2pt}
\newcommand{\ci}[1]{{\scriptsize[#1]}}
\begin{tabular*}{\linewidth}{@{\extracolsep{\fill}}lcccccc@{}}
\toprule
Arm & Orders & Unav.\ (\%) & Entries
& \textsc{Accurate} (\%) & $\Delta$ (pp) & $\tau$ (\%) \\
\midrule
Released (free) & 8.5 & 35.0 & 16.7
& \shortstack{46.5\\\ci{35.2, 57.7}} & ---
& \shortstack{28.9\\\ci{22.1, 35.9}} \\
\addlinespace
Budget, 10 tests & 10 & 25.1 & 10.0
& \shortstack{43.7\\\ci{32.4, 54.9}} & \shortstack{$-2.8$\\\ci{$-12.7$, 7.0}}
& \shortstack{25.9\\\ci{20.1, 31.9}} \\
\addlinespace
Budget, single entry & 10 & 29.2 & 6.5
& \shortstack{36.6\\\ci{25.4, 47.9}} & \shortstack{$-9.9$\\\ci{$-19.7$, 0.0}}
& \shortstack{20.9\\\ci{16.2, 26.0}} \\
\addlinespace
Checklist, 10 & 10 & 3.8 & 12.4
& \shortstack{42.3\\\ci{31.0, 53.5}} & \shortstack{$-4.2$\\\ci{$-14.1$, 5.6}}
& \shortstack{22.9\\\ci{18.0, 28.0}} \\
\addlinespace
Checklist, 15 & 15 & 8.5 & 20.8
& \shortstack{40.8\\\ci{29.6, 52.1}} & \shortstack{$-5.6$\\\ci{$-15.5$, 4.2}}
& \shortstack{31.9\\\ci{26.4, 37.4}} \\
\addlinespace
Atomic, named & 10 & 14.1 & 8.6
& \shortstack{32.4\\\ci{21.1, 43.7}} & \shortstack{$-14.1$\\\ci{$-23.9$, $-4.2$}}
& \shortstack{14.6\\\ci{10.3, 19.2}} \\
\addlinespace
Atomic, matched & 10 & 5.2 & 9.5
& \shortstack{35.2\\\ci{23.9, 46.5}} & \shortstack{$-11.3$\\\ci{$-21.1$, $-1.4$}}
& \shortstack{15.9\\\ci{12.1, 19.9}} \\
\addlinespace
Random, 10 & 10 & 34.9 & 8.0
& \shortstack{25.4\\\ci{15.5, 35.2}} & \shortstack{$-21.1$\\\ci{$-33.8$, $-8.5$}}
& \shortstack{7.3\\\ci{4.3, 10.5}} \\
\bottomrule
\end{tabular*}
\end{table}

\paragraph{Effect of an atomic-test budget.} The itemised budget reduces the entries released from $16.7$ per clip over $8.5$ orders to $10.0$ distinct entries over $10$ orders ($532$ of $710$ orders answered, $1.53$ entries per answered order before deduplication), while $\tau$ is $25.9\%$ against $28.9\%$ in the released arm (Table~\ref{tab:app-selection}). Accuracy is similar: $43.7\%$ with the atomic budget and $46.5\%$ in the released arm. The observed decisive-evidence yield persists when requests are itemised.

\paragraph{Comparison of ten-item strategies.} The model-selected budget, fixed checklist and random list each issue ten items and reach $43.7\%$, $42.3\%$ and $25.4\%$ accuracy, with $\tau$ of $25.9\%$, $22.9\%$ and $7.3\%$ (Table~\ref{tab:app-selection}). Model selection and the checklist exceed random selection by $18.3$ and $16.9$ points but differ from each other by only $1.4$ points in accuracy. The fifteen-item checklist raises $\tau$ to $31.9\%$ while accuracy is $40.8\%$. Availability differs across arms: $3.8\%$ of checklist orders return nothing, versus $25.1\%$ of model-selected and $34.9\%$ of random orders. The checklist returns more results, and its decisive-evidence coverage is close to model selection at the ten-item budget (a $3.0$-point difference whose interval includes zero).

\paragraph{Comparison at a matched ten-order budget.} The ten-item arms above match list length but release $10.0$, $12.4$ and $8.0$ entries per clip. We therefore limit each order to at most one chart entry for both the model-selected list and two checklist variants. The named checklist returns nothing for $14.1\%$ of orders, partly because full blood count has no entry in $38$ of $71$ charts; the matched variant returns nothing for $5.2\%$. Under the symmetric release rule, the model-selected list reaches $36.6\%$ accuracy and $20.9\%$ $\tau$, versus $35.2\%$ and $15.9\%$ for the matched checklist (Table~\ref{tab:app-selection}). Neither accuracy nor decisive-evidence coverage separates the lists clearly ($5.0$ points of $\tau$ in favour of model selection, interval including zero). Restricting the model's own orders to one entry reduces its accuracy by $7.0$ points and $\tau$ by $5.0$ points, showing that multi-entry release contributes to coverage. The original checklist's decisive hits came mostly from the eye-movement, contrast-MRI and glucose items. Blinded raters see the same ten orders in both model arms and give similar overall scores to the structured lists, while counting more unnecessary orders in the model's list (Table~\ref{tab:app-rating}).

\paragraph{Blinded rating of the order lists.} Two blinded raters scored the five ten-item arms on all $71$ clips, seeing the video, visible sign, history exchange and one list at a time, without the chart, reference diagnosis or other arms (Table~\ref{tab:app-rating}). Both place the random list last ($1.35$ and $1.85$ overall) and the four structured lists together at $2.7$--$2.8$. They count $3.4$--$3.6$ unnecessary orders per ten in the model-selected list, versus $2.6$--$3.1$ in the checklists and $5.8$--$7.7$ in the random list. Across $355$ rated lists, accuracy is higher when a decisive test is obtained ($58.3\%$ vs.\ $13.9\%$; $n{=}175$ and $180$). Like the raters, $\tau$ separates random from structured lists, and a decisive test tracks accuracy.

\begin{table}[htbp]
\centering
\caption{\textbf{Blinded rating of the ten-item order lists} on all $71$ clips.
Paired values report Rater~1 / Rater~2; scores and unnecessary-order counts
are means; $\tau$ is computed on the same clips.}
\label{tab:app-rating}
\small
\setlength{\tabcolsep}{4pt}
\begin{tabular*}{\linewidth}{@{\extracolsep{\fill}}lccc@{}}
\toprule
Strategy & Overall score (1--5) & Orders judged unnecessary & $\tau$ (\%) \\
\midrule
Random, 10         & 1.35 / 1.85 & 7.7 / 5.8 &  7.3 \\
Checklist, 10      & 2.82 / 2.67 & 2.6 / 2.9 & 22.9 \\
Atomic, named      & 2.80 / 2.79 & 2.6 / 3.1 &  14.6 \\
Atomic, matched    & 2.76 / 2.77 & 2.8 / 3.1 &  15.9 \\
Model-selected, 10 & 2.82 / 2.69 & 3.4 / 3.6 & 25.9 \\
\bottomrule
\end{tabular*}
\end{table}

\FloatBarrier
\section{Temporal-Window Training: A Paired Analysis}
\label{app:window}

Post-training improves the small describer; where its window is placed matters little, but a short, densely sampled segment appears more useful than the same budget spread over the clip (Section~\ref{sec:temporal_results}). Below we give the teacher's windows, the adapter versus the untrained model, a blinded rating of the sign sentences, paired window-minus-random contrasts, and a sign-only adapter trained without windows. Each clip has three downstream runs per arm; comparisons use clip-level means and $10{,}000$ paired bootstrap resamples.

\paragraph{Window statistics.} The teacher was run on all $71$ clips and returned a usable record for $68$; it judged the $32$-frame survey sufficient for $29$ of these. The implied sampling rate within a window is $0.7$, $1.4$ and $2.9$ frames per second at $K=8,16,32$.

\paragraph{The untrained model on identical inputs.} Table~\ref{tab:base-vs-adapter} evaluates unadapted Qwen3.5-4B on the same survey-plus-window input as the adapted student. Unadapted accuracy is $22$--$27\%$ across budgets. The adapted student leads by $5.2$, $8.0$ and $6.6$ points at $K=8,16,32$, and its coverage is higher by $3.9$, $5.2$ and $5.1$ points, with paired intervals excluding zero at every budget; accuracy point estimates also favour the adapter.

\paragraph{Independent recognition of the student's sentences.}
A blinded judge compared the sentences produced by the window-trained student and by the
unadapted model against the reference phenomenology, component by component
(Table~\ref{tab:window-signs}). The adapted student named the body part correctly in $66\%$ of
clips (unadapted: $40\%$) and the movement character in $54\%$ (unadapted: $10\%$), and its
sentence was judged the closer one in $48$ clips against $14$, with $6$ ties. Laterality is rarely
correct in either model ($9\%$ versus $5\%$ of the $57$ clips whose reference states a side), and
the activation condition is named correctly less often after training ($18\%$ versus $38\%$).
Sign recognition therefore improves with training on the two components the training target
emphasises, assessed independently of the consultation.

\paragraph{Paired contrasts for window position.}
The paired differences (Table~\ref{tab:window-paired}) remain small across $K=8,16,32$: accuracy changes by $-0.5$ to $+3.3$ points and source-workup coverage by at most $1.6$ points. The data thus support the adapter effect more clearly than the choice of window.

\paragraph{A sign-only adapter without windows.}
To separate post-training from the window, we trained a second adapter with the same $68$ clips, source-grouped folds, target sentence and prompt, for one epoch per fold, on $32{+}K$ frames spaced evenly over the whole clip, so that no window enters training or inference. Table~\ref{tab:no-window} compares it with the window-trained adapter at the same frame budget. Its coverage matches the window-trained adapter at every budget ($18.3$--$19.0\%$ versus $17.3$--$19.4\%$) and exceeds the untrained model by $4.7$--$5.6$ points. Its accuracy ($26.8$--$27.2\%$) stays within $5.2$ points of the untrained model and trails the window arm by $2.3$, $2.8$ and $6.6$ points at $K=8,16,32$; the random-window arm also leads it at every budget, by $2.8$, $1.4$ and $3.3$ points. Learning to describe the sign therefore accounts for the coverage gain, whereas the accuracy lead of both windowed arms suggests that a short, densely sampled segment is more useful than the same budget spread over the clip; these differences remain within sampling error.

\begin{table}[H]
\centering
\caption{\textbf{Paired window-minus-random differences} (percentage points; 95\% paired case-level bootstrap intervals, 71 clips). W: window arm; R: random window
of the same length. Both arms share the adapter and the $32$-frame survey, as in
Table~\ref{tab:downstream diagnosis}.}
\label{tab:window-paired}
\small
\setlength{\tabcolsep}{5pt}
\begin{tabular}{lcc}
\toprule
$K$ & $\Delta$ accuracy (pp) & $\Delta\tau$ (pp) \\
\midrule
8  & $-0.5$ $[-7.5, +6.6]$ & $+0.9$ $[-2.2, +4.0]$ \\
16 & $+1.4$ $[-7.0, +9.4]$ & $+0.7$ $[-1.9, +3.3]$ \\
32 & $+3.3$ $[-4.2, +10.8]$ & $-1.6$ $[-5.0, +1.8]$ \\
\bottomrule
\end{tabular}
\end{table}

\begin{table}[H]
\centering
\caption{\textbf{Blinded judgement of the student's sign sentences}, $K{=}32$ with the
$32$-frame survey. For each clip the judge saw the reference phenomenology and the two
sentences and marked each component independently; unstated components are not scored, so denominators differ.}
\label{tab:window-signs}
\small
\setlength{\tabcolsep}{4pt}
\begin{tabular}{lccccc}
\toprule
sentence & body part & character & side & condition & judged closer (clips) \\
\midrule
no adapter             & 27/68 (40\%) & 7/68 (10\%)  & 3/57 (5\%) & 13/34 (38\%) & 14 \\
window-trained adapter & 45/68 (66\%) & 37/68 (54\%) & 5/57 (9\%) & 6/34 (18\%)  & 48 \\
\bottomrule
\end{tabular}
\end{table}

\begin{table}[H]
\centering
\caption{\textbf{Untrained Qwen3.5-4B versus the window-trained student} on the same survey $+$ window
input ($32$-frame survey, same day, denominator $71$, three runs per clip, clip-level
bootstrap). Coverage is source-workup coverage ($\tau$); differences are paired over clips.}
\label{tab:base-vs-adapter}
\small
\setlength{\tabcolsep}{3pt}
\begin{tabular*}{\linewidth}{@{\extracolsep{\fill}}lcccccc@{}}
\toprule
 & \multicolumn{2}{c}{untrained} & \multicolumn{2}{c}{adapter} & \multicolumn{2}{c}{adapter $-$ untrained} \\
\cmidrule(lr){2-3}\cmidrule(lr){4-5}\cmidrule(lr){6-7}
$K$ & acc.\ [95\% CI] & cov. & acc.\ [95\% CI] & cov. & $\Delta$acc.\ [95\% CI] & $\Delta$cov.\ [95\% CI] \\
\midrule
8  & 23.9 [16.4, 31.9] & 13.4 & 29.1 [20.7, 38.0] & 17.3 & $+5.2$ $[-3.8, +14.1]$ & $\mathbf{+3.9}$ $[+0.1, +7.9]$ \\
16 & 22.1 [14.6, 30.5] &  13.4 & 30.0 [21.1, 39.4] & 18.6 & $+8.0$ $[-1.4, +17.4]$ & $\mathbf{+5.2}$ $[+1.5, +9.2]$ \\
32 & 27.2 [18.3, 36.6] & 14.3 & 33.8 [24.4, 43.7] & 19.4 & $+6.6$ $[-1.9, +15.5]$ & $\mathbf{+5.1}$ $[+0.6, +9.8]$ \\
\bottomrule
\end{tabular*}
\end{table}

\begin{table}[H]
\centering
\caption{\textbf{Sign-only adapter trained without windows} versus the window-trained adapter at matched frame budget (denominator $71$, three runs per clip, clip-level paired bootstrap). No window: adapter trained and evaluated on $32{+}K$ evenly spaced frames. Window / random: the window-trained adapter on the $32$-frame survey plus $K$ frames from the student's predicted window or from a random window of the same length. Untrained: Qwen3.5-4B on the survey-plus-window input. Coverage is source-workup coverage ($\tau$); differences are no window minus the listed arm.}
\label{tab:no-window}
\small
\setlength{\tabcolsep}{3pt}
\begin{tabular*}{\linewidth}{@{\extracolsep{\fill}}llcccc@{}}
\toprule
$K$ & arm & acc.\ [95\% CI] & cov. & $\Delta$acc.\ [95\% CI] & $\Delta$cov.\ [95\% CI] \\
\midrule
8  & no window & 26.8 [19.2, 34.3] & 19.0 & -- & -- \\
   & window    & 29.1 [20.7, 38.0] & 17.3 & $-2.3$ $[-10.8, +5.6]$ & $+1.8$ $[-1.0, +4.6]$ \\
   & random    & 29.6 [21.6, 38.5] & 16.4 & $-2.8$ $[-10.8, +4.7]$ & $+2.7$ $[-0.3, +5.8]$ \\
   & untrained & 23.9 [16.4, 31.9] & 13.4 & $+2.8$ $[-5.2, +10.3]$ & $\mathbf{+5.6}$ $[+2.1, +9.5]$ \\
\midrule
16 & no window & 27.2 [19.2, 35.7] & 18.3 & -- & -- \\
   & window    & 30.0 [21.1, 39.4] & 18.6 & $-2.8$ $[-11.7, +6.1]$ & $-0.3$ $[-3.1, +2.4]$ \\
   & random    & 28.6 [20.2, 37.6] & 17.9 & $-1.4$ $[-8.9, +6.1]$ & $+0.3$ $[-2.6, +3.3]$ \\
   & untrained & 22.1 [14.6, 30.5] & 13.4 & $+5.2$ $[-2.3, +12.7]$ & $\mathbf{+4.9}$ $[+1.0, +8.9]$ \\
\midrule
32 & no window & 27.2 [19.2, 35.7] & 18.9 & -- & -- \\
   & window    & 33.8 [24.4, 43.7] & 19.4 & $-6.6$ $[-13.6, +0.5]$ & $-0.4$ $[-3.7, +2.9]$ \\
   & random    & 30.5 [21.6, 39.9] & 20.9 & $-3.3$ $[-10.3, +3.8]$ & $-2.0$ $[-5.4, +1.4]$ \\
   & untrained & 27.2 [18.3, 36.6] & 14.3 & $0.0$ $[-8.0, +8.5]$ & $+4.7$ $[-0.5, +9.8]$ \\
\bottomrule
\end{tabular*}
\end{table}

\FloatBarrier
\section{Recognition across Frame Budgets and Clinical Categories}
\label{app:frames}

Two analyses illustrate Section~\ref{sec:visual_results}: recognizing a sign rarely places its cause in the differential, and in a worked example more frames do not make the differential more focused.

\subsection{A Frame-Budget Worked Example}

\begin{figure}[ht]
\centering
\includegraphics[width=\linewidth]{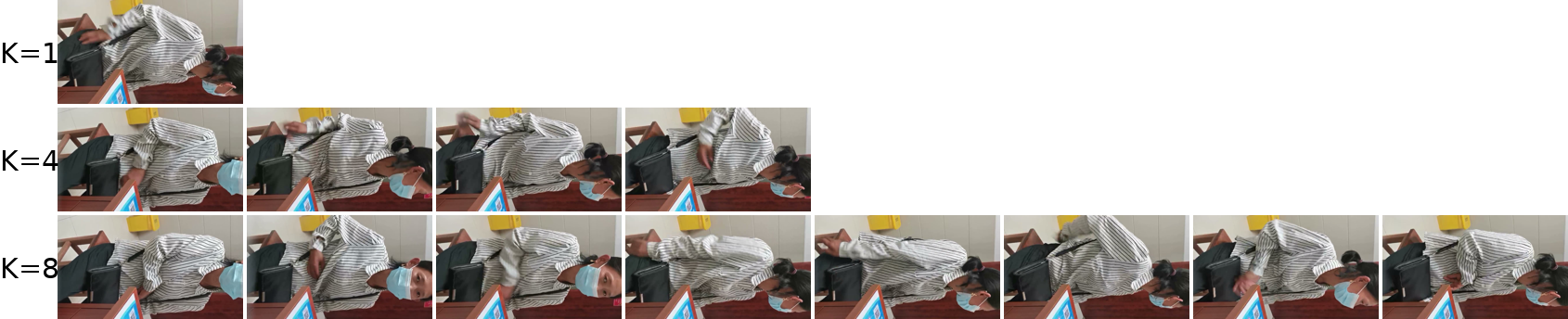}
\caption{\textbf{One clip at three frame budgets.} Uniformly sampled frames from the same $17.8$-second clip at
$K{=}1$, $4$ and $8$. Frames share a common scale; each row
shows the full sampling budget.}
\label{fig:app-frames}
\end{figure}

We trace how the frame budget changes GPT-5.6-luna's description
and initial differential for PMC9815763, a woman with right-arm
chorea from Graves' hyperthyroidism. The reference phenomenology
is \emph{continuous, rapid, irregular movements of one arm}; the reference
aetiology is hyperthyroidism. Figure~\ref{fig:app-frames} shows the input at
three budgets, and Table~\ref{tab:app-frames} compares responses at six.

\begin{table}[ht]
\centering
\small
\caption{\textbf{Responses at six frame budgets.} Phenomenology recognition and differential diagnosis, quoted from the responses.
\emph{Sign grade} is the grader's rating of phenomenology recognition.}
\label{tab:app-frames}
\begin{tabular}{@{}c>{\raggedright\arraybackslash}p{0.44\linewidth}>{\raggedright\arraybackslash}p{0.30\linewidth}c@{}}
\toprule
$K$ & Phenomenology recognition & Differential diagnosis & Sign grade \\
\midrule
$1$ & ``lying on their side \ldots{} a prominent dark/purplish area with swelling around the
      ear/temple and upper cheek'' & acute trauma to the head; periorbital or auricular haematoma
      & wrong \\
$2$ & ``repeatedly bend forward and hold/rub the upper abdomen or lower chest'' & upper
      abdominal pain; gastritis, ulcer or biliary disease & wrong \\
$4$ & ``markedly bent forward/stooped \ldots{} intermittent, somewhat irregular/flinging or jerky
      movement of the opposite arm'' & camptocormia from Parkinson disease or another parkinsonian
      syndrome & partial \\
$8$ & ``the movements look irregular, flowing, and somewhat twisting rather than rhythmic'' &
      hyperkinetic movement disorder; chorea or choreoathetosis, possibly hemichorea & correct \\
$32$ & ``irregular, seemingly involuntary movements \ldots{} jerky and variable, with prominent
       upper-limb and whole-body involvement'' & functional movement disorder; ``the marked
       variability \ldots{} support[s] this over a fixed neurologic syndrome'' & correct \\
$128$ & ``recurrent, forceful, involuntary movements of the trunk and upper limbs \ldots{}
        episodic and somewhat irregular'' & functional movement disorder, particularly functional
        truncal jerks or dystonia & partial \\
\bottomrule
\end{tabular}
\end{table}

\paragraph{Recognition across frame budgets.} At $K{=}1$, the model
interprets a static appearance as swelling; at $K{=}2$, it interprets the
postural change as pain. At $K{=}4$, it notices irregular arm movement but
emphasises posture, receiving partial credit. At $K{=}8$, it describes
irregular, flowing, non-rhythmic movement. The $K{=}32$ answer also receives a correct grade; at $K{=}128$ the description misses the continuous, one-sided character and is graded partial. 

\paragraph{Recognition does not necessarily narrow the differential.}
From $K{=}8$ onward, the model lists five to eight candidates spanning
functional, drug-induced, seizure-related and movement-syndrome
explanations. At higher budgets, functional movement disorder often
leads the list despite correct or partial phenomenology grades, with variability
cited in support. In this example, correct sign recognition can coexist
with a broad differential that additional frames do not consistently narrow.

\subsection{Recognition and Aetiological Coverage by Category}
\label{app:lines}

\begin{figure}[H]
\centering
\includegraphics[width=0.655\linewidth]{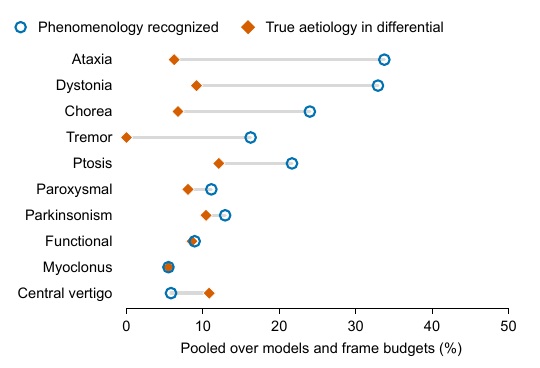}
\caption{\textbf{Recognition and aetiological coverage by category.} Phenomenology recognition and aetiological coverage
pooled over five models and eight frame budgets and ordered by their
difference. Facial palsy is excluded because it contains only one clip.}
\label{fig:lines}
\end{figure}

Figure~\ref{fig:lines} compares phenomenology recognition with aetiological
coverage by clinical category, pooling five models and eight frame
budgets; categories are ordered by the gap between them.

Recognition exceeds coverage most clearly in ataxia ($33.8\%$ versus $6.2\%$), dystonia ($32.9\%$ versus $9.2\%$), chorea ($24.0\%$ versus $6.8\%$) and tremor ($16.2\%$ versus $0.0\%$). In these categories naming a visible sign does not translate into a compatible aetiological hypothesis, consistent with recognition and hypothesis formation being distinct hurdles (Section~\ref{sec:visual_results}).

The two measures are close in paroxysmal events ($11.1\%$ versus $8.1\%$), parkinsonism ($12.9\%$ versus $10.4\%$), functional movement disorder ($8.9\%$ versus $8.6\%$) and myoclonus ($5.5\%$ for both). Only central vertigo reverses the pattern ($5.8\%$ recognition versus $10.8\%$ coverage), where a broad differential can include a compatible candidate despite an inaccurate description. Neither measure tracks the other across categories, so the two are evaluated separately.

\FloatBarrier
\section{Retrieval: Candidate Coverage and a Consultation Example}
\label{app:lit}

\begin{figure}[!htbp]
\centering
\includegraphics[width=\linewidth]{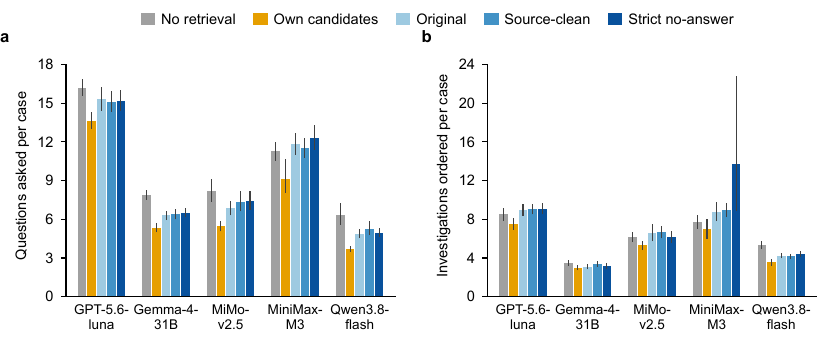}
\caption{\textbf{Interaction counts under retrieval conditions} (71 cases): (\textbf{a})~questions asked and
(\textbf{b})~investigations ordered per case. Colours denote no retrieval (grey), Own candidates (orange), Original (light blue), Source-clean (blue),
and Strict no-answer (dark blue). Bars show case-level means with 95\% bootstrap intervals.}
\label{fig:retrieval-interactions}
\end{figure}

This appendix supports Section~\ref{sec:retrieval_results}. Retrieval supplies candidate causes the model misses, decontamination removes source articles without lowering mean accuracy, the work-up moves toward the documented one for four of five systems, and an equal-length list of unrelated causes reproduces much of the accuracy effect. A worked consultation closes the section. Figure~\ref{fig:retrieval-interactions} reports the questions asked and investigations ordered per case under each condition.

\begin{figure}[ht]
\centering
\includegraphics[width=\linewidth]{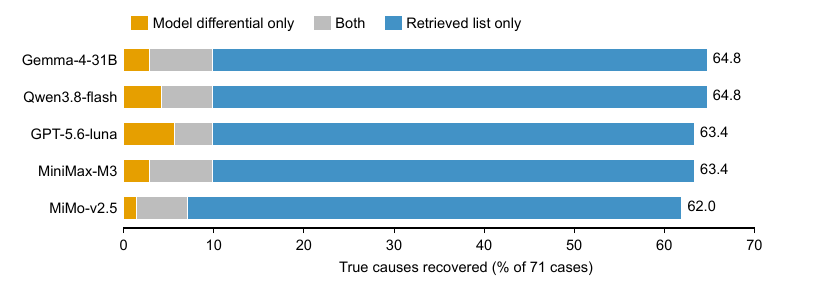}
\caption{\textbf{The retrieved list covers causes the model misses.} Coverage of reference causes by the model's initial differential
and the Source-clean retrieved list. Stacked segments distinguish causes covered only
by the model, by both lists, or only by retrieval; the total is their union.}
\label{fig:union}
\end{figure}

\paragraph{Complementarity of candidate lists.}
The retrieved list recovers the true cause far more often than the model's video-only differential, and the two cover partly different cases. Across five models the retrieved list covers $57.7$--$62.0\%$ of clips and the model alone $7.0$--$9.9\%$; their union covers $62.0$--$64.8\%$ (Figure~\ref{fig:union}). The model's own hypotheses still add $1.4$--$5.6$ points that retrieval misses, so we supply the combined list rather than replacing the model's hypotheses with retrieved candidates.

\begin{table}[htbp]
\centering
\caption{\textbf{Retrieval conditions minus the video condition} on the same $71$ clips (percentage points, $95\%$ case-level bootstrap intervals). Accuracy and coverage $\tau$ as in Table~\ref{tab:consultation_conditions}.}
\label{tab:retrieval-paired}
\footnotesize
\setlength{\tabcolsep}{3pt}
\begin{tabular*}{\linewidth}{@{\extracolsep{\fill}}llcc@{}}
\toprule
System & Condition & $\Delta$ accuracy [95\% CI] & $\Delta\tau$ [95\% CI] \\
\midrule
GPT-5.6-luna & Source-clean & $+2.8$ $[-9.9, +15.5]$ & $+8.3$ $[+0.9, +15.6]$ \\
             & Strict no-answer & $+1.4$ $[-11.3, +14.1]$ & $+5.3$ $[-3.0, +13.4]$ \\
Gemma-4-31B  & Source-clean & $-2.8$ $[-14.1, +8.5]$ & $+0.7$ $[-5.7, +6.5]$ \\
             & Strict no-answer & $-4.2$ $[-15.5, +7.0]$ & $-2.0$ $[-8.3, +3.9]$ \\
MiMo-v2.5    & Source-clean & $+21.1$ $[+8.5, +33.8]$ & $+13.3$ $[+5.0, +21.3]$ \\
             & Strict no-answer & $+15.5$ $[+2.8, +28.2]$ & $+11.0$ $[+4.7, +17.3]$ \\
MiniMax-M3   & Source-clean & $+15.5$ $[+1.4, +28.2]$ & $+8.0$ $[+2.0, +14.4]$ \\
             & Strict no-answer & $+23.9$ $[+9.9, +36.6]$ & $+13.6$ $[+6.7, +20.7]$ \\
Qwen3.8-flash & Source-clean & $+11.3$ $[0.0, +23.9]$ & $+9.0$ $[+2.0, +15.6]$ \\
             & Strict no-answer & $+11.3$ $[0.0, +21.1]$ & $+7.0$ $[+0.3, +13.7]$ \\
\bottomrule
\end{tabular*}
\end{table}

\paragraph{Decontamination.}
Source-clean filtering removed the source article for $24/71$ cases; shared-DOI and near-duplicate-title rules removed no additional records. Strict no-answer filtering removed $999$ records across $41$ cases whose titles or excerpts name the confirmed diagnosis or an accepted equivalent. Table~\ref{tab:filtering} gives candidate coverage and accuracy under the three filtering levels. Source-clean filtering lowers the coverage of the supplied list from $68.4$ to $63.7\%$ and changes accuracy by $-4.2$ to $+7.0$ points; Strict no-answer filtering lowers it by a further $13.3$ points, to $50.4\%$.

We audited the $31{,}749$ distinct records retained after source-clean filtering for other reports of the same patient. Of these, $144$ records in $35$ cases share at least one author with the source article; the $43$ that also share two or more diagnosis words describe other patients or reviews from the same groups. An author-independent check found $7$ records whose abstract describes a patient of the case's age and sex and shares a rare diagnosis term; each differs from the source patient in country, year or presentation, and no record meets both criteria. The audit found no same-patient report among the retained abstracts; reports identifiable only from full text remain possible.

\begin{table}[htbp]
\centering
\caption{\textbf{Candidate coverage and accuracy under the three filtering levels} ($71$ cases). Coverage is the share of cases whose supplied list (the model's own Stage~1 differential followed by the retrieved causes) contains the true cause, graded with the Appendix~\ref{app:prompts} rank probe; accuracy is the \textsc{Accurate} rate of the consultation.}
\label{tab:filtering}
\small
\begin{tabular}{lcccccc}
\toprule
 & \multicolumn{3}{c}{candidate coverage (\%)} & \multicolumn{3}{c}{accuracy (\%)} \\
\cmidrule(lr){2-4}\cmidrule(lr){5-7}
system & original & source-clean & strict & original & source-clean & strict \\
\midrule
GPT-5.6-luna & 70.4 & 63.4 & 52.1 & 50.7 & 49.3 & 47.9 \\
Gemma-4-31B & 70.4 & 64.8 & 52.1 & 29.6 & 31.0 & 29.6 \\
MiMo-v2.5 & 64.8 & 62.0 & 47.9 & 36.6 & 43.7 & 38.0 \\
MiniMax-M3 & 69.0 & 63.4 & 47.9 & 45.1 & 40.8 & 49.3 \\
Qwen3.8-flash & 67.6 & 64.8 & 52.1 & 33.8 & 33.8 & 33.8 \\
\midrule
Mean & 68.4 & 63.7 & 50.4 & 39.2 & 39.7 & 39.7 \\
\bottomrule
\end{tabular}
\end{table}

\paragraph{Paired differences from the video condition.} Source-clean retrieval raises $\tau$ with an interval above zero for four of five systems, and accuracy for MiMo-v2.5 and MiniMax-M3; the Qwen3.8-flash accuracy interval reaches zero (Table~\ref{tab:retrieval-paired}). Under the strict audit, $\tau$ still rises with an interval above zero for three systems, and the accuracy gains of MiMo-v2.5 and MiniMax-M3 remain.

\paragraph{Blinded order-list ratings.}
Table~\ref{tab:retrieval-joint} adds the blinded ratings of Appendix~\ref{app:tau} to the accuracy and source-workup outcomes. For MiMo-v2.5, MiniMax-M3, and Qwen3.8-flash, retrieval improves these measures together; unnecessary-order shares also fall for MiMo-v2.5 and Qwen3.8-flash. The matched-list control below tests how much of the gain depends on retrieved content.

\begin{table}[t]
\centering
\caption{\textbf{Retrieval under the full set of measures} ($71$ cases). Accuracy and $\tau$ as
in Table~\ref{tab:consultation_conditions}; orders is the mean per case. Quality ($1$--$5$) and
unnecessary share (unnecessary orders over orders placed) are blinded ratings of each order list,
reported as Rater~1 / Rater~2. Own: the model's own Stage~1 differential without retrieval;
clean: the same list followed by source-clean retrieved causes.}
\label{tab:retrieval-joint}
\footnotesize
\setlength{\tabcolsep}{3pt}
\begin{tabular*}{\linewidth}{@{\extracolsep{\fill}}llccccc@{}}
\toprule
system & arm & acc.\ (\%) & $\tau$ (\%) & orders & quality & unnecessary (\%) \\
\midrule
GPT-5.6-luna  & video & 46.5 & 28.9 & 8.5 & 3.38 / 3.10 & 22 / 25 \\
              & own   & 52.1 & 37.9 & 7.5 & 3.56 / 3.20 & 19 / 24 \\
              & clean & 49.3 & 37.2 & 9.1 & 3.35 / 3.22 & 24 / 25 \\
\addlinespace
Gemma-4-31B   & video & 33.8 & 20.3 & 3.5 & 2.67 / 2.46 & 24 / 31 \\
              & own   & 18.3 & 12.0 & 3.0 & 2.43 / 2.46 & 32 / 38 \\
              & clean & 31.0 & 20.9 & 3.4 & 2.76 / 2.52 & 20 / 36 \\
\addlinespace
MiMo-v2.5     & video & 22.5 & 16.3 & 6.2 & 2.63 / 2.46 & 37 / 42 \\
              & own   & 28.2 & 15.3 & 5.3 & 2.72 / 2.49 & 32 / 41 \\
              & clean & 43.7 & 29.6 & 6.7 & 3.00 / 2.89 & 24 / 26 \\
\addlinespace
MiniMax-M3    & video & 25.4 & 19.3 & 7.7 & 2.51 / 2.41 & 38 / 44 \\
              & own   & 18.3 & 17.9 & 7.0 & 2.44 / 2.31 & 46 / 48 \\
              & clean & 40.8 & 27.2 & 9.0 & 2.61 / 2.69 & 40 / 37 \\
\addlinespace
Qwen3.8-flash & video & 22.5 & 16.3 & 5.3 & 2.58 / 2.32 & 32 / 42 \\
              & own   & 18.3 & 12.6 & 3.6 & 2.69 / 2.68 & 27 / 33 \\
              & clean & 33.8 & 25.2 & 4.2 & 3.34 / 2.94 & 15 / 27 \\
\bottomrule
\end{tabular*}
\end{table}

\paragraph{Length- and form-matched control.}
The mismatched condition appends causes retrieved for a donor case from another category. A fixed per-clip seed and truncation to the source-clean list length match the form and number of candidates in all $71$ cases. Own candidates alone lowers accuracy below the video condition for Gemma-4-31B, MiniMax-M3 and Qwen3.8-flash, and appending mismatched causes raises it again; for GPT-5.6-luna neither appended list changes accuracy reliably. Table~\ref{tab:retrieval-mismatch} shows a content-specific accuracy gain for MiMo-v2.5, and higher $\tau$ with source-clean than with mismatched candidates for MiMo-v2.5 and Qwen3.8-flash, with paired intervals excluding zero; unrelated candidates reproduce part of the accuracy gains in MiniMax-M3 and Qwen3.8-flash.

\begin{table}[t]
\centering
\caption{\textbf{Length- and form-matched control for retrieval} ($71$ cases). Cells give
accuracy / $\tau$ (\%). Own: the model's own Stage~1 differential; mismatched: the same list
followed by source-clean causes retrieved for an unrelated case, truncated to the length of the
source-clean list; clean: source-clean retrieval. Differences are in accuracy percentage points
with $95\%$ paired case-level bootstrap intervals. The paired $\tau$ difference, clean $-$ mismatched, is $-0.7$ $[-7.4, +6.2]$ for GPT-5.6-luna, $+3.7$ $[-3.9, +11.0]$ for Gemma-4-31B, $+13.0$ $[+6.7, +19.6]$ for MiMo-v2.5, $+4.3$ $[-1.6, +10.2]$ for MiniMax-M3 and $+9.3$ $[+4.3, +14.5]$ for Qwen3.8-flash.}
\label{tab:retrieval-mismatch}
\footnotesize
\setlength{\tabcolsep}{2pt}
\begin{tabular*}{\linewidth}{@{\extracolsep{\fill}}lcccccc@{}}
\toprule
system & video & own & mismatched & clean & clean $-$ mismatched & mismatched $-$ own \\
\midrule
GPT-5.6-luna  & 46.5 / 28.9 & 52.1 / 37.9 & 53.5 / 37.9 & 49.3 / 37.2 & $-4.2$ $[-15.5, +7.0]$ & $+1.4$ $[-7.0, +9.9]$ \\
Gemma-4-31B   & 33.8 / 20.3 & 18.3 / 12.0 & 22.5 / 17.3 & 31.0 / 20.9 & $+8.5$ $[-2.8, +19.7]$ & $+4.2$ $[-5.6, +14.1]$ \\
MiMo-v2.5     & 22.5 / 16.3 & 28.2 / 15.3 & 22.5 / 16.6 & 43.7 / 29.6 & $+21.1$ $[+9.9, +32.4]$ & $-5.6$ $[-16.9, +5.6]$ \\
MiniMax-M3    & 25.4 / 19.3 & 18.3 / 17.9 & 29.6 / 22.9 & 40.8 / 27.2 & $+11.3$ $[-1.4, +23.9]$ & $+11.3$ $[0.0, +22.5]$ \\
Qwen3.8-flash & 22.5 / 16.3 & 18.3 / 12.6 & 31.0 / 15.9 & 33.8 / 25.2 & $+2.8$ $[-7.0, +12.7]$ & $+12.7$ $[+2.8, +22.5]$ \\
\bottomrule
\end{tabular*}
\end{table}

\paragraph{Consultation example.} A shorter, more targeted order list can recover more decisive evidence. In the illustrative myoclonus case, both runs ask $18$ history questions. With retrieval, the model orders $12$ rather than $18$ investigations, receives $53$ rather than $36$ chart entries, and obtains two of four decisive entries rather than none. It drops cultures, toxicology, serum ammonia, and head CT while adding nerve-conduction studies with electromyography and a formal oculomotor assessment.

In the retrieval condition, the returned decisive findings include
rhythmic EMG bursts synchronised with the visible jerks and an EEG
report documenting no EEG--EMG correlation. The final answer is
\emph{post-infectious subcortical myoclonus with cerebellar ataxia},
graded \textsc{ACCURATE}; without retrieval, it is
\emph{functional neurological symptom disorder}, graded
\textsc{NOT ACCURATE}.

\FloatBarrier
\section{Robustness to Corrupted History Responses}
\label{app:lie}

Histories can be wrong, including withheld or misreported information \citep{levy2018prevalence}. We flip a fraction of the structured history responses, keep the video and questions fixed, and measure the change in investigation selection and final diagnosis. At $80\%$ corruption accuracy falls for every model, and more orders do not compensate.

\subsection{Experimental Protocol}
\label{sec:res-lie}

For each consultation, we retain the video and history questions
from the uncorrupted run and flip a specified fraction of the
structured responses (\texttt{yes} to \texttt{no}; \texttt{no} or \texttt{unknown} to \texttt{yes}). The model then orders investigations
and diagnoses from the modified history. Requested results
come from the unchanged source-case chart.

Fixing the questions isolates the downstream response to altered history and precludes follow-up questions that might resolve inconsistencies. Each corruption level is compared with its uncorrupted baseline on accuracy, investigation count and $\tau$.

\subsection{Aggregate Results}

\begin{figure}[t]
\centering
\includegraphics[width=\linewidth]{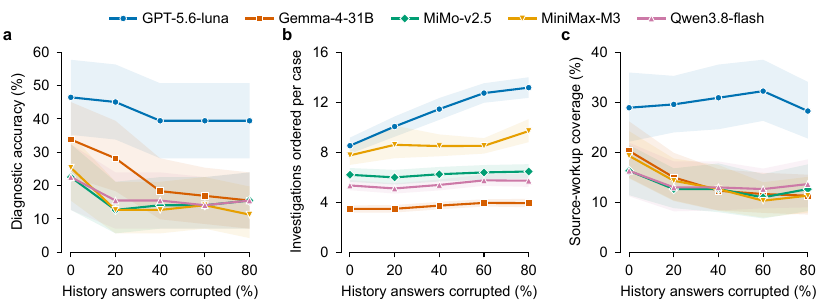}
\caption{\textbf{Consultation outcomes across history-corruption levels}: (\textbf{a})~diagnostic accuracy,
(\textbf{b})~investigation count, and (\textbf{c})~source-workup coverage. The video and history
questions are held fixed; investigation orders and final diagnoses are generated using the modified
responses. Shaded bands are marginal 95\% bootstrap intervals over clips.}
\label{fig:lie}
\end{figure}

At $80\%$ corruption, diagnostic accuracy is lower than the
uncorrupted baseline in all five models
(Figure~\ref{fig:lie}). Gemma-4-31B and MiniMax-M3 decline by
$18.3$ and $14.1$ percentage points, respectively, alongside
reductions in source-workup coverage.

Changes in workup size vary across models. GPT-5.6-luna increases
its mean investigation count from $8.5$ to $13.2$ while maintaining
approximately stable source-workup coverage. MiniMax-M3 shows a
smaller increase in investigation count but lower diagnostic
accuracy. Additional orders therefore do not consistently
compensate for corrupted history.

\subsection{Worked Example: Neuropathic Tremor in Charcot--Marie--Tooth Disease}
\label{sec:app-worked-example}

In case \texttt{mild\_neuropathic\_CMT\_PMC11600610}, the video shows postural tremor associated with CMT2A \citep{mohammed2024cmt2a}. Four chart entries are decisive: limb power and wasting, pes cavus, nerve conduction studies with electromyography (NCS/EMG), and targeted genetic testing. The source report describes one patient.

The model asks the same $18$ questions in both runs. At nominal $40\%$ corruption, seven replies change: one \texttt{yes} becomes \texttt{no}, one \texttt{no} becomes \texttt{yes}, and five \texttt{unknown} become \texttt{yes}. Tables~\ref{tab:app-lie-q} and~\ref{tab:app-lie-prov} give the complete question-level responses and source evidence; a compound question returns \texttt{unknown} unless the documented evidence settles the whole statement.

The altered history points the model toward an acute cerebral work-up. It drops NCS/EMG, its only decisive hit without corruption, and requests stroke imaging, EEG, and cardiac monitoring. Both runs place $14$ orders, but returned chart entries fall from $29$ to $19$ and decisive-entry recovery from $1/4$ to $0/4$. The diagnosis changes from hereditary axonal CMT2 (\textsc{ACCURATE} at the disease-entity level) to functional movement disorder (\textsc{NOT ACCURATE}).

\begin{table}[htbp]
\centering
\footnotesize
\caption{\textbf{History questions and source-aligned responses} in the worked example
(a 34-year-old man, the sole patient of the source case report).
\emph{Basis} keys into Table~\ref{tab:app-lie-prov}, which quotes the
supporting source evidence. A dash means no passage is quoted here, not
verified absence. Question~8 is supported by S6.
Starred rows are corrupted at the nominal $40\%$ rate.}
\label{tab:app-lie-q}
\begin{tabularx}{\linewidth}{@{}r>{\raggedright\arraybackslash}Xccc@{}}
\toprule
 & Question & Basis & Uncorr. & Corr. \\
\midrule
 1$^{*}$ & Are you experiencing involuntary shaking, jerking, or flapping
      movements? & S1 & yes & \textbf{no} \\
 2$^{*}$ & Did these movements begin suddenly?
      & --- & unknown & \textbf{yes} \\
 3$^{*}$ & Are the movements present when your hands are at rest?
      & S2 & no & \textbf{yes} \\
 4 & Do the movements worsen when you hold your arms out or use your
      hands? & S2 & yes & yes \\
 5 & Do both hands move in the same way? & --- & unknown & unknown \\
 6$^{*}$ & Have you noticed weakness or clumsiness on one side of your body?
      & --- & unknown & \textbf{yes} \\
 7 & Have you had numbness or loss of sensation? & S3 & yes & yes \\
 8 & Have you had difficulty speaking, understanding speech, seeing, or
      walking? & S6 & yes & yes \\
 9 & Have you had a severe or unusual headache? & --- & unknown & unknown \\
10 & Have you lost awareness or had episodes of staring or unresponsiveness?
      & --- & unknown & unknown \\
11$^{*}$ & Have you recently had confusion, unusual sleepiness, or a change
      in your alertness? & --- & unknown & \textbf{yes} \\
12 & Have you had fever, vomiting, or significant dehydration recently?
      & --- & unknown & unknown \\
13 & Do you have known liver disease? & S4 & no & no \\
14 & Do you have known kidney disease? & S4 & no & no \\
15 & Do you have diabetes or thyroid disease?
      & S4$^{\dagger}$ & unknown & unknown \\
16 & Have you started, stopped, or changed any medications recently?
      & --- & unknown & unknown \\
17$^{*}$ & Do you use alcohol or recreational drugs?
      & S5$^{\dagger}$ & unknown & \textbf{yes} \\
18$^{*}$ & Have you had similar episodes before?
      & --- & unknown & \textbf{yes} \\
\bottomrule
\end{tabularx}

\vspace{2pt}
\footnotesize
$^{\dagger}$ Settled in part only: S4 excludes diabetes but the report does
not address thyroid disease, and S5 excludes illicit drug use but does not
quantify alcohol use. Under the compound-question rule of
Appendix~\ref{app:prompts-env} the environment returns \texttt{unknown}.
\end{table}

\begin{table}[htbp]
\centering
\footnotesize
\caption{\textbf{Patient-level provenance} for every non-\texttt{unknown} response in
the worked example. Sentences quote the single-patient source report
\citep{mohammed2024cmt2a} (\texttt{PMC11600610}, \emph{Journal of Medical
Case Reports}, CC BY-NC-ND 4.0);
\emph{Loc.} gives its section.}
\label{tab:app-lie-prov}
\begin{tabularx}{\linewidth}{@{}c>{\raggedright\arraybackslash}p{0.26\linewidth}
    >{\raggedright\arraybackslash}Xl@{}}
\toprule
Key & Chart entry & Sentence in the source & Loc. \\
\midrule
S1 & \texttt{hand tremor when holding a posture: yes};
     \texttt{the tremor started about six months ago: yes}
   & ``[He] came to our hospital with a complaint of tremors of the hands of
     6 months duration.'' --- ``The tremor was limited to the hands.''
   & Case pres. \\
\addlinespace
S2 & \texttt{tremor at rest: no}
   & ``We noted a postural tremor that attenuated when the patient assumed
     an anatomic position.''
   & Case pres. \\
\addlinespace
S3 & \texttt{reduced sensation in the feet: yes}
   & ``His sensory examination revealed a mildly decreased vibratory
     sensation; otherwise, he had no sensory level.''
   & Clin. findings \\
\addlinespace
S4 & \texttt{known liver disease: no};
     \texttt{known kidney disease: no}; \texttt{diabetes: no}
   & ``[T]he patient had no history of diabetes, hypertension, psychiatric,
     cardiac, renal, liver diseases, or cancers.''
   & Case pres. \\
\addlinespace
S5 & \texttt{illicit drug use: no}
   & ``He had no history of illicit drug use and denied constitutional
     symptoms.''
   & Case pres. \\
\addlinespace
S6 & \texttt{difficulty walking: yes}
   & Walking difficulty and bilateral distal weakness are documented.
   & Patient info. \\
\bottomrule
\end{tabularx}
\end{table}

\clearpage
\section{Prompts}
\label{app:prompts}

The benchmark prompts are reproduced verbatim, with runtime substitutions
shown as angle-bracketed placeholders. They are organised by function:
model evaluation (\S\ref{app:prompts-model}), responses from the patient and
chart environment (\S\ref{app:prompts-env}), retrieval
(\S\ref{app:prompts-retrieval}), and grading (\S\ref{app:prompts-grading}). Prompts to the model under evaluation run on that model, and the teacher prompt of
Appendix~\ref{app:temporal_training_details} on \texttt{openai/gpt-5.6-luna}; every other prompt, including the verifier
used in Appendix~\ref{app:tau}, runs on \texttt{deepseek/deepseek-v4.1-flash} through OpenRouter.

\subsection{The Model under Evaluation}
\label{app:prompts-model}

\paragraph{Stage 1, from the video alone.} Sent with $K$ frames as interleaved images.

{\scriptsize
\begin{verbatim}
You are a doctor. Below are <K> frame(s) sampled in order across a ~<duration>
second video of a patient.

Describe what you see, then give your primary guess of the disease. You may list
several possible diagnoses.
\end{verbatim}
}

\paragraph{Stage 2, turn 1: the history.} The opening clause carries the role under test
(\emph{You are a neurologist seeing a new patient} or \emph{You are a doctor seeing a new
patient}); all reported runs use the latter. The candidate block is present only in the retrieval
conditions and their list controls.

{\scriptsize
\begin{verbatim}
<role> Above are <K> frames sampled across a <duration>-second video of the
consultation, in order.

<candidate block, retrieval conditions only>You may now take a history, but under one
constraint: **Ask yes/no questions. The environment returns yes, no, or unknown
when the record does not establish the answer**,
and the patient will answer only yes, no, or unknown. Anything you do not ask, you do not
learn.

Ask as many or as few questions as you judge this patient needs. Stop when you
believe another question would not change what you think is wrong. Put them in the
order you would ask them - most informative first.

One question per line, numbered. Nothing else.
\end{verbatim}
}

In the Blind condition the frames are omitted and the first paragraph is replaced by:

{\scriptsize
\begin{verbatim}
You are a doctor. A new patient has been referred to you and you have not yet seen
or examined them. You know nothing about them at all.
\end{verbatim}
}

The candidate block, when present, precedes the constraint:

{\scriptsize
\begin{verbatim}
A literature search on the signs visible in this video returned the following
reported causes. The list is not guaranteed to contain this patient's cause, and
most entries in it are wrong for this patient:

<candidate causes>
\end{verbatim}
}

\paragraph{Stage 2, turn 2: the investigations.}

{\scriptsize
\begin{verbatim}
The patient answered:

<numbered questions, each with yes, no, or unknown>

You may now investigate. There is no fixed list to choose from.
Name whatever you would actually order, in your own words:
bedside examination manoeuvres, blood tests, imaging,
electrophysiology, invasive procedures, or a therapeutic trial.

A therapeutic trial is returned only if you name that specific
trial. Asking to "try treatment" returns nothing.

Order as many or as few investigations as you judge this patient
needs. Stop when another test would not change what you think is
wrong. Put the investigations most likely to settle the diagnosis
first.

One investigation per line, numbered. Name the test, not what you
expect it to show. Nothing else.
\end{verbatim}
}

\paragraph{Stage 2, turn 3: the diagnosis.}

{\scriptsize
\begin{verbatim}
The results are:

<one block per order: the order as written, then the chart entries it covered with
their values, or "not performed / not available">

For orders marked "not performed / not available", no documented
result is available in this environment. Do not infer a result.

Now give your diagnosis. State the single diagnosis you believe is correct - the
disease entity and its cause, as specifically as the evidence allows - then, on
separate lines, up to three alternatives you would still consider.

Begin with "DIAGNOSIS:" followed by the single answer on one line.
\end{verbatim}
}

\subsection{The Environment}
\label{app:prompts-env}

Both prompts below align free text with the case record.
The patient prompt returns yes/no/unknown responses based only on
documented features. The chart prompt returns matching entry names,
and the harness retrieves their recorded results.
Neither prompt is permitted to invent clinical findings.

\paragraph{The patient.} The environment answers each question from the
documented symptom table: \texttt{yes} for an established finding,
\texttt{no} for an explicitly absent finding, and \texttt{unknown} when
the record is insufficient. Omission from the table indicates unreported
information, not a negative finding.

{\scriptsize
\begin{verbatim}
A doctor asked a patient these yes/no questions:

<numbered questions>

The patient's documented features are:

<the case's symptom table>

Answer each question using only the documented features above.

Use exactly one of the following responses:

- "yes" if the documented features explicitly establish that the complete
  statement is true;
- "no" if the documented features explicitly establish that the complete
  statement is false;
- "unknown" if the documented features do not provide enough information
  to determine whether the complete statement is true or false.

Do not infer unreported features. Absence from the documented feature table
does not imply absence of the clinical finding and must not by itself produce
a "no" response.

For compound questions, return yes or no only when the documented
evidence determines the truth of the complete statement.
Otherwise, return unknown.

Reply with ONLY a JSON object:

{"<question number>": "yes"|"no"|"unknown", ...}

\end{verbatim}
}

\paragraph{The chart.}

{\scriptsize
\begin{verbatim}
A doctor ordered these investigations. A numbered line may bundle several tests.

<numbered orders>

This chart holds results for exactly these entries:

<the case's investigation menu, therapeutic trials marked [NAMED-ONLY]>

For each numbered order, list the chart entries it covers - matching on what the
test is, not on wording. An order covers an entry only if it genuinely asks for
that test. Entries marked [NAMED-ONLY] are therapeutic trials: list one only if
that specific trial is named, never for a general request to try treatment.

Reply with ONLY a JSON object mapping each order number to a list of chart entry
names, exactly as written above: {"1": ["..."], "2": [], ...}
\end{verbatim}
}

\subsection{Retrieval}
\label{app:prompts-retrieval}

\paragraph{Filtering.}
For Source-clean and Strict no-answer retrieval, before candidate-cause
extraction, we remove records matching the
source PMCID or DOI, or having title-token Jaccard similarity
$\geq 0.85$ with the source title. Source metadata are fetched from
Europe PMC by PMCID. Titles are lower-cased, stripped of
non-alphanumeric characters, and tokenised, dropping a fixed
list of generic publication terms and function words.

Strict no-answer retrieval additionally checks titles and the first $260$
abstract characters for the confirmed diagnosis or case-specific
accepted equivalents. The diagnosis is truncated at the first dash
to exclude supporting reasoning. Matching uses the same normalisation,
discards equivalent strings shorter than four characters, and removes
records containing any remaining equivalent as a substring.
Because it uses ground truth, it serves only as a leakage control, not as a
deployable filter.

\paragraph{Normalisation.} The model's description is converted to canonical
phenomenology terms before HPO matching. This vocabulary-normalisation
step has no access to the diagnosis.

{\scriptsize
\begin{verbatim}
Rewrite the clinical description below as a list of standard neurological
phenomenology terms, the kind used as headings in a movement-disorder textbook.

Rules:
- one term per line, nothing else, no numbering, no explanation
- use the canonical noun form: "Gait ataxia", "Chorea", "Resting tremor",
  "Ptosis", "Nystagmus", "Bradykinesia", "Dystonia", "Myoclonus", "Facial palsy",
  "Spasticity"
- include the body region as a separate term when the description gives one
- when the description states a side or a distribution, add it on its own line
  using exactly one of: Unilateral, Bilateral, Left, Right, Generalized, Focal,
  Multifocal, Axial, Proximal, Distal, Alternating laterality
- include only what the description actually states; do not infer a diagnosis
- if the description states no abnormality, output the single line NONE

Description:
<the model's own description of the sign>
\end{verbatim}
}

\paragraph{Extraction.} Applied to the titles and abstract openings returned by the Europe PMC
query.

{\scriptsize
\begin{verbatim}
Below are published papers, all retrieved because they describe the same clinical
sign. Each entry is a title followed by the start of its abstract.

For each entry, name the underlying CAUSE or DISEASE it was about, in four words
or fewer. If the paper is a review that enumerates several causes, list them
separated by " ; ". One entry per line, in the same order, nothing else. Write
SKIP if no cause is named.

<titles and abstract openings>
\end{verbatim}
}

\subsection{Grading}
\label{app:prompts-grading}

\paragraph{Stage 1: the sign and the differential.}

{\scriptsize
\begin{verbatim}
You are grading a vision model that was shown a short clinical video of one
patient and asked, with NO history and NO test results, to describe what it saw
and propose possible diagnoses.

Evaluate two separate outcomes:
(1) recognition of the visible phenomenology;
(2) coverage of a diagnostic hypothesis compatible with the reference case.

Diagnostic coverage does NOT require exact identification of the confirmed
disease. A clinically appropriate syndrome or broader aetiological category
may qualify under the case-specific acceptance criteria below.

GROUND TRUTH
  confirmed diagnosis : <true diagnosis>
  reference visible phenomenology : <reference sign>
  additional hypotheses accepted as COVERAGE, including qualifying syndromes
  and broader aetiological categories : <accept-as-coverage>
  related but insufficient hypotheses, NOT accepted as coverage :
    <related-but-not-covered>

THE MODEL'S ANSWER (free prose)
<the answer>

Read the whole answer. Grade SIGN and DIAGNOSIS independently.
Correct sign recognition does not by itself establish diagnostic coverage,
and an incorrect sign description does not rule out diagnostic coverage.

SIGN - did it identify the visible abnormality?
  correct : identifies the reference phenomenology or an equivalent physical
            description anywhere in the answer; clinical synonyms count.
  partial : describes a relevant abnormality in the correct body region,
            but the description is incomplete or too vague to establish
            the reference phenomenology.
  wrong   : identifies an incompatible phenomenology or the wrong body
            region, or does not describe a relevant visible abnormality.

DIAGNOSIS - does a proposed hypothesis cover the reference case?
  correct : identifies the confirmed disease or an equivalent diagnosis,
            OR identifies a syndrome or broader aetiological category
            included in the case-specific COVERAGE list.
            Equivalent clinical terminology is accepted.
            Exact disease identification and equal specificity to the
            confirmed diagnosis are NOT required for accepted hypotheses.
  partial : proposes a clinically related hypothesis that does not satisfy
            the COVERAGE criteria, including an entry in the
            related-but-not-covered list.
  wrong   : proposes an incompatible or unrelated hypothesis, or does not
            propose a diagnostic hypothesis.

Apply these rules:
- Do not reject an accepted hypothesis merely because it is broader than
  the confirmed disease.
- Do not automatically accept every broad syndrome or disease category.
  Broad hypotheses must match the case-specific COVERAGE list or an
  equivalent clinical expression.
- Merely repeating the visible sign does not establish coverage unless
  that expression also names an explicitly accepted diagnostic syndrome.
- Count hypotheses the model proposes as possibilities, even if they are
  not its leading diagnosis.
- Do not count a diagnosis mentioned only to deny or explicitly rule it out.
- Do not infer a diagnostic hypothesis that the model did not express.

Assign dx1 to the explicitly designated primary diagnosis, if present;
otherwise use the first proposed diagnosis. Assign dx2 and dx3 to the next
two distinct proposed diagnoses in presentation order.
Use "wrong" for missing slots.

Assign best using ALL distinct proposed diagnoses in the complete answer,
including diagnoses beyond dx3:
  correct : at least one hypothesis is graded correct.
  partial : none is correct, but at least one is partial.
  wrong   : all are wrong, or no diagnostic hypothesis is proposed.

The binary aetiological-coverage outcome is 1 if best is "correct",
and 0 otherwise. Partial hypotheses do not receive coverage credit.

Reply with ONLY this JSON:
{"sign":"correct|partial|wrong",
 "dx1":"correct|partial|wrong",
 "dx2":"correct|partial|wrong",
 "dx3":"correct|partial|wrong",
 "best":"correct|partial|wrong",
 "reason":"<= 12 words"}
\end{verbatim}
}

\paragraph{Stage 1: the uncapped rank probe.}
This probe reads the complete answer and records the number of distinct
proposed diagnoses and the rank of the first hypothesis satisfying the
case-specific aetiological-coverage criteria in
Section~\ref{sec:measurements}.
Accepted syndromes and broader aetiological categories qualify;
exact identification of the confirmed disease is not required.

{\scriptsize
\begin{verbatim}
A model was shown frames from a video of one patient and asked to describe
what it saw and propose possible diagnoses, with no history or test results.

TRUE DIAGNOSIS: <true diagnosis>
Additional hypotheses accepted as COVERAGE, including qualifying
syndromes and broader aetiological categories: <accept-as-coverage>
Related but insufficient hypotheses, NOT accepted as coverage:
<related-but-not-covered>

THE MODEL'S ANSWER:
<the answer>

Read the whole answer and list, in order of first appearance, every
distinct diagnosis proposed as a possibility. Merge synonymous entries
and exclude diagnoses mentioned only to rule them out.

Return the rank of the first hypothesis that covers the reference case:
either the confirmed disease or an explicitly accepted syndrome or
broader aetiological category. Equivalent clinical terminology counts.
Exact disease identification and equal specificity to the confirmed
diagnosis are NOT required. Do not accept other broad hypotheses unless
equivalent to an explicitly accepted hypothesis. Merely repeating the
visible sign does not qualify unless it names an accepted syndrome.

n_listed is the total number of distinct proposed diagnoses.
rank is the 1-based position of the first qualifying hypothesis.
Use rank = 0 if none qualifies. Consider the entire list, not only
the first three entries.

Reply with ONLY {"n_listed": <int>, "rank": <int>}
\end{verbatim}
}

\paragraph{Stage 2: the committed diagnosis.} Only the primary diagnosis is graded; the
alternatives the model is allowed to append are recorded but not credited.
A grade of \texttt{none} means \textsc{Not accurate}.

{\scriptsize
\begin{verbatim}
You are grading a doctor who watched a video of a patient, took a yes/no history,
ordered investigations, and then named a diagnosis.

GROUND TRUTH
  true diagnosis : <true diagnosis>
  count as accurate (the actual disease entity) : <accept-as-accurate>
  Accepted as partial: <accept-as-partial>

THE DOCTOR'S PRIMARY DIAGNOSIS
<the diagnosis>

Grade the primary diagnosis alone.
  accurate     : matches a case-specific accepted correct diagnosis,
               including an accepted broader disease or syndrome
               formulation; equivalent clinical terminology counts

partial      : matches the case-specific partial-credit criteria,
               including a related syndrome, cause, or broad disease
               family, but does not satisfy the correct-answer criteria

none         : satisfies neither the correct-answer criteria nor
               the partial-credit criteria

Reply with ONLY {"grade":"accurate|partial|none","reason":"<= 12 words"}
\end{verbatim}
}

\clearpage
\section{Temporal-Window Training: Reproducibility Details}
\label{app:temporal_training_details}

\subsection{Teacher: Offline Window Supervision}
\label{app:teacher}

The frozen teacher generates offline targets for temporal localisation.
Given the video and two reference clinical fields, it selects at most one
window or returns a no-window decision. Only the resulting text target is
used to train the student; the teacher is absent at inference.

\paragraph{Model and decoding.} We use \texttt{openai/gpt-5.6-luna} through
OpenRouter, pinned to the OpenAI provider.
The teacher runs once per clip (temperature $0$, top-$p$ $0.9$,
output limit $1100$ tokens).

\paragraph{Inputs.} Each silent clip is represented by $32$ uniformly
sampled frames, $380$\,px wide with preserved aspect ratio and JPEG quality
$85$. Each frame carries a top-left timestamp over an opaque patch,
measured in seconds from clip start to two decimals. Images precede text
in one user turn. The text supplies frame count, survey rate
($32/\mathrm{duration}$), inter-frame interval, native recording rate,
and shot boundaries for multi-shot clips.

The two clinical inputs are the verbatim reference phenomenology sentence
and confirmed diagnosis. The prompt distinguishes windows located from
visible frames, the reference sentence, or inference from the diagnosis.
The teacher's explanations are not supplied to the student.

\paragraph{Target construction.} We parse the last JSON object in the
teacher's reply and convert a selected interval to its start time and span
(end minus start), in seconds. The student target is text containing only
a mode and, for a window request, one start/span pair. Frame rates, shot
identifiers, confidence scores and explanations are excluded.

A \texttt{survey\_sufficient=yes} response produces a \texttt{sufficient}
target, ignoring any optional window. An \texttt{obtainable=no} response
produces an \texttt{unobtainable} target. Contradictory flags are treated
as invalid rather than as window requests.

\paragraph{Separation of supervision and input.} Targets are generated
offline before the five-fold split; each student is trained only on the
targets in its training folds. Held-out teacher outputs are excluded.
Neither student pass receives clinical context during training or inference.
The reference phenomenology sentence serves only as the recognition target.

\paragraph{Full teacher prompt.} The versioned template below is filled
with video metadata, shot boundaries, and the two clinical fields.
Doubled braces are literal-brace escapes for string substitution.

\clearpage
\noindent\textbf{Teacher prompt (single-window protocol).}
\begingroup
\fontsize{6.6}{8.5}\selectfont
\begin{verbatim}
Above are {n_frames} frames covering the whole {duration}-second clip, in order, with the time
stamped on each. That is {sample_fps} frames per second - {sample_ms} milliseconds pass between one
frame and the next. The recording itself runs at {native_fps} frames per second.{shots_block}

You already know what this clip contains:
  what a clinician sees: {sign}
  what the patient has:  {diagnosis}

That sentence and that diagnosis are given to you and NOT to the person your window is for. Keep
track of which of the two you are using at each step: some of what you know is visible in these
frames, and some of it is not.

Your job is NOT to diagnose. It is to say which seconds of this recording, shown at what frame
rate, would let a viewer who knows nothing about it see the abnormality for themselves.

You choose only WHICH SECONDS and HOW MANY FRAMES PER SECOND. You cannot ask for a crop, a zoom, a
closer view, or any change of framing - the whole frame will be shown as it stands, at the size you
see above. If the thing is too small on screen to read at this framing, no choice of seconds or
rate will fix that, and the honest answer is that it cannot be obtained.

Nothing you ask for will be fetched. You are not being tested on whether you can then see it; you
are being asked to state the best request you can make, and to be explicit about how you arrived
at it.

Work in this order, and let the later steps follow from the earlier ones rather than being chosen
first.

1. What kind of thing is it.
     static_appearance   a feature you could read off a single still - a posture held, a lid
                         position, a pupil, an asymmetry that does not change
     continuous_movement something that goes on throughout, so any long enough stretch contains it
     episodic_event      something that happens at particular moments and not between them
     task_evoked         something that appears only while the patient is doing something, or while
                         being examined

2. Which part of the body, and how large it is on screen. If the part appears more than once in the
   picture - two hands, two eyes, two legs - say which one, by the side of the IMAGE it is on, not
   by the patient's own left and right. Give roughly what fraction of the frame width it occupies,
   and judge honestly whether a change in that part is readable at this size.

3. How long ONE occurrence lasts, in milliseconds - not how often it comes back. These are
   different numbers and only the first sets the frame rate: a movement lasting eighty milliseconds
   that returns every half second needs frames close enough together to catch the eighty, not the
   five hundred. Give both if the thing repeats.

   For a viewer to see an occurrence at all it must fall on at least two frames, so the frames must
   be closer together than HALF its duration:

     required frames per second  =  2000 / (duration of one occurrence in ms)

   Work that number out and compare it with the {sample_fps} frames per second above.

4. Propose at most ONE window: a stretch of seconds and a frame rate, entirely inside ONE shot.
   The windows array must contain zero or one entry, never more than one.

   For the window, say what it rests on:
     read_from_frames        you can point to the stamped frames where it is visible
     stated_in_the_sentence  the sentence you were given already names the moment or the task -
                             "on outstretching the arms", "while walking", "as the patient keeps
                             talking" - and your window follows from those words rather than from
                             anything you saw
     inferred_from_condition neither: you cannot see it here and the sentence does not say when,
                             so you are reasoning about when it is likely to be happening - while
                             the part is being used, while it is being tested, when it is at rest,
                             whatever the condition implies

   All three are legitimate. Answer stated_in_the_sentence whenever it is true, even if the frames
   happen to confirm it; we are counting how often the sentence, rather than the recording, is what
   located the sign.

   Be careful of one trap. A thing fast enough to need a higher rate is, by that very fact, a thing
   these frames are too slow to show, so you cannot expect to watch it happen and point there - and
   the movement that IS plainly visible above is the slow one, which is usually not the thing that
   matters. Do not simply name the seconds where the most conspicuous movement is.

   Give the window a confidence between 0 and 1, meaning how likely you think it is that a viewer
   shown that window, and told nothing, would describe the abnormality correctly.

   Two answers stand outside the window request, and you should give them when they are true:
     survey_sufficient   the frames above already show it; no window is needed
     obtainable = no     no choice of seconds or rate will show it at this framing

   You may still give one optional window when survey_sufficient is yes, but say so in the flag.
   If obtainable is no, return an empty windows array.

Separately, list the stamped times at which the thing can actually be SEEN in the frames above -
the ones you would point to as evidence. If it cannot be made out in any of them, give an empty
list. A run of consecutive frames is not evidence unless each of them shows it.

Your reasons will be read by a human reviewer and never shown to a model being trained. Even so,
write them so that they could be: name the part, the side, the size on screen, the timing, the
direction, and nothing that identifies a condition.

Reply as JSON:
{{"kind": "static_appearance or continuous_movement or episodic_event or task_evoked",
  "body_part": "...",
  "image_side": "left or right or either or not_applicable",
  "size_fraction_of_width": NUMBER between 0 and 1,
  "readable_at_this_size": "yes or no",
  "occurrence_duration_ms": NUMBER or null,
  "repetition_interval_ms": NUMBER or null,
  "required_fps": NUMBER or null,
  "sentence_names_when": "yes or no",
  "survey_sufficient": "yes or no",
  "obtainable": "yes or no",
  "not_obtainable_reason": "too_small or not_in_recording or other or null",
  "windows": [
    {{"shot": SHOT_NUMBER, "at_s": [START_SECONDS, END_SECONDS], "fps": NUMBER,
      "basis": "read_from_frames or stated_in_the_sentence or inferred_from_condition",
      "confidence": NUMBER between 0 and 1,
      "why": "one sentence"}}
  ],
  "evidence_s": [SECONDS, ...],
  "reason": "two or three sentences, following from steps 1 to 3"}}
Nothing else.
\end{verbatim}
\endgroup

\clearpage
\subsection{Student: Training and Two-Pass Inference}
\label{app:student}

The student shares one Qwen3.5-4B model and one LoRA adapter across two
tasks: predicting a window from the survey and describing the visible
phenomenology, both as plain text generation.

\paragraph{Two-pass inference.} Pass 1 receives the timestamped
$32$-frame survey, without clinical fields, and predicts at most one
start/span pair. Pass 2 generates one phenomenology sentence. A valid window
adds an image--instruction turn containing $K$ sampled frames; otherwise,
recognition reuses the existing survey without presenting it again. The
rules below determine which frames are used. Frame selection is discrete
and does not propagate gradients.

\begin{center}\small
\begin{tabularx}{\linewidth}{@{}>{\raggedright\arraybackslash}p{0.35\linewidth}>{\raggedright\arraybackslash}X@{}}
\toprule
Pass-1 response & Recognition evidence \\
\midrule
Valid window & $K\in\{8,16,32\}$ evenly sampled frames within the window, without timestamps \\
\texttt{sufficient} or \texttt{unobtainable} & Original $32$ survey frames; no new frames or repeated presentation \\
Invalid JSON & Skip window sampling; use the original survey without presenting it again \\
\bottomrule
\end{tabularx}
\end{center}

\paragraph{Text targets and instructions.} Localisation predicts the
teacher-derived mode and start/span, excluding reasoning and frame rate.
For illustration, the interval $[2.0,3.5]$ seconds can be represented as
\texttt{\{"mode":"request","start":2.0,"span":1.5\}}.
At $K=16$, recognition receives $16$ evenly sampled frames from that
interval. The instruction requests at most one
within-shot window, or a no-window response when the survey suffices or the
sign cannot be captured at the available framing. It requests no diagnosis,
explanation, crop, zoom, or frame rate.

Recognition uses this instruction ($N=K$ for a valid window,
$N=32$ for a fallback):
\begingroup
\fontsize{7.2}{9}\selectfont
\begin{verbatim}
These {N} frames are sampled from a silent patient clip. In one sentence,
say what a clinician would see - the abnormality itself, in plain physical
words. Name no disease.
\end{verbatim}
\endgroup
Its training target is the benchmark clinician's phenomenology sentence,
verbatim. Neither pass receives that sentence, the diagnosis, or other
clinical context as input.

\paragraph{Objective.} Let $z_{1:L}$ be the tokenized teacher-derived window
text and $y_{1:T}$ the reference phenomenology sentence. Both losses are
mean autoregressive token cross-entropies:
\[
\mathcal{L}_{\mathrm{win}}
= -\frac{1}{L}\sum_{\ell=1}^{L}
\log p_\theta(z_\ell\mid z_{<\ell},V_{32},I_{\mathrm{win}}),
\qquad
\mathcal{L}_{\mathrm{phen}}
= -\frac{1}{T}\sum_{t=1}^{T}
\log p_\theta(y_t\mid y_{<t},V_N,I_{\mathrm{phen}}).
\]
\[
\mathcal{L}=0.3\,\mathcal{L}_{\mathrm{win}}
+0.7\,\mathcal{L}_{\mathrm{phen}}.
\]
Here $V_{32}$ is the survey, $V_N$ is the pass-2 input, and $I$ is
the instruction. Only target-text tokens contribute to
loss; video, instruction, and template-prefix positions are masked.
Only adapter parameters $\theta$ are trained, with no temporal-bin
classifier, coordinate-regression head or frame-rate loss.

\clearpage
\paragraph{Training schedule and parameters.} Recognition uses
teacher-derived windows in epoch 1. During epoch 2, the probability of using
the student's own prediction increases from $0$ to $0.5$. No-window
responses and invalid JSON trigger the same survey fallback as at inference.
Five-fold cross-validation groups cases by source article, with each
student trained on four folds and evaluated on the held-out fold. The loss
weight is fixed across folds. The table below lists all settings.
\begin{center}\small
\begin{tabularx}{\linewidth}{@{}>{\raggedright\arraybackslash}p{0.23\linewidth}>{\raggedright\arraybackslash}X@{}}
\toprule
Base model & Qwen3.5-4B\\
Adapted & language side only; vision tower frozen \\
LoRA & $r=16$, $\alpha=32$, dropout $0.05$, no bias \\
Target modules & \texttt{q,k,v,o,gate,up,down\_proj}, \texttt{in\_proj\_qkv}, \texttt{out\_proj}; \texttt{visual} excluded \\
Trainable & $27.9$M of $4.57$B ($0.61\%$) \\
\midrule
Optimiser & AdamW, weight decay $0$ \\
Learning rate & $1\times10^{-4}$, one-cycle, warmup $10\%$ of steps \\
Batch & $1$ sample, gradient accumulation $4$, gradient clipping $1.0$ \\
Precision & bfloat16, gradient checkpointing \\
Epochs (window adapter) & $1$ teacher-forced, $1$ with scheduled sampling \\
\midrule
Survey (pass 1) & $32$ frames, $380$\,px wide, JPEG $q{=}85$, timestamp burnt in \\
Recognition (pass 2) & valid window: $K\in\{8,16,32\}$, $384$\,px wide, no timestamp; otherwise reuse original $32$-frame survey \\
Loss weight & $\lambda=0.3$ \\
Scheduled sampling & student window with probability ramped $0\to0.5$ over epoch $2$ \\
\midrule
Folds & $5$, grouped by source article, fixed once and reused across budgets \\
Template & chat template rendered with the reasoning block closed \\
\midrule
Teacher output limit & $1100$ tokens; parsing takes the last JSON object in the reply \\
Pass 1 decoding & greedy (\texttt{do\_sample=false}), at most $1100$ new tokens \\
Pass 2 decoding & greedy (\texttt{do\_sample=false}), at most $64$ new tokens \\
\midrule
Python / PyTorch & 3.10 / 2.8.0 (CUDA 12.8 runtime) \\
Transformers / PEFT & \texttt{5.13.0.dev0} / 0.19.1 \\
GPU / driver & NVIDIA RTX A6000 (48\,GB), driver \texttt{535.309.01} \\
\bottomrule
\end{tabularx}
\end{center}

\paragraph{Frame budget and comparison.} A valid window adds $K$ frames to the $32$-frame survey; a fallback reuses the survey without further frames. Short windows may repeat source frames when $K$ exceeds the number available at the native recording rate. Random preserves the predicted window duration but changes its position within the clip; both arms use the same adapter and fallback rule. The sign-only adapter uses the same clips, folds, target sentence and prompt, but is trained for one epoch per fold on $32{+}K$ evenly spaced frames. Paired comparisons appear in Appendix~\ref{app:window}.

\clearpage
\section{Grading and Order-Matching Audits}
\label{app:grader}

We check three aspects of reliability: agreement between
automatic and human grading (\S\ref{app:human-grading}),
order-to-chart matching (\S\ref{app:matcher}), and the retrieval contrast
under independent human grading (\S\ref{app:sym-grading}).

\subsection{Alignment with Human Grading}
\label{app:human-grading}

An independent human reviewer re-graded sampled responses using the same
rubric and blinding.

\paragraph{Sampling.}
For Stage~1, the sampling unit is one response per system, frame budget, and
clip; we drew $100$ responses uniformly at random from this population using a
fixed seed. Four lacked usable paired grades and were excluded, leaving $96$
for analysis. For Stage~2, no sampling was needed: the human reviewer graded
the final diagnosis of every video-condition consultation of GPT-5.6-luna, one
response per clip, giving $71$ paired grades. Both lists are released with
the benchmark.

\paragraph{Agreement.}
We report exact grade agreement and quadratically weighted Cohen's $\kappa_w$.
Exact agreement means both graders assign the same category
(not necessarily \textsc{Accurate}).

\begin{table}[htbp]
\centering
\small
\caption{\textbf{Agreement between the model grader and the independent human
reviewer}; Stage~2 evaluation is restricted to the video condition (GPT-5.6-luna, all $71$
consultations).}
\label{tab:grader-agreement}
\begin{tabular}{@{}llrrr@{}}
\toprule
Stage & Outcome & $n$ & Agreement & $\kappa_w$ \\
\midrule
1 & Phenomenology recognition & $96$ & $89.6\%$ & $0.903$ \\
  & Aetiology                 & $96$ & $96.9\%$ & $0.921$ \\
\midrule
2 & Final diagnosis           & $71$ & $91.5\%$ & $0.947$ \\
\bottomrule
\end{tabular}
\end{table}

For the primary Stage~2 outcome the graders agree on $65$ of $71$ responses
(Table~\ref{tab:grader-confusion}); all six disagreements involve adjacent grades, and none
spans two levels.

\begin{table}[htbp]
\centering
\small
\caption{\textbf{Paired Stage~2 grades} in the video condition ($71$ consultations).
Rows represent the model grader and columns the human reviewer.}
\label{tab:grader-confusion}
\begin{tabular}{lrrr}
\toprule
Model / Human & Accurate & Partial & Not accurate \\
\midrule
Accurate     & $32$ & $1$  & $0$  \\
Partial      & $1$  & $10$ & $2$  \\
Not accurate & $0$  & $2$  & $23$ \\
\bottomrule
\end{tabular}
\end{table}

\subsection{Human Verification of Order Matching}
\label{app:matcher}

We sampled $100$ investigation orders from the video condition across five
systems, uniformly at random with a fixed seed. The sample spans $52$ clips.
A human reviewer compared each order with its returned findings and the
complete chart menu. Of these orders, $95$ were matched correctly, $3$
returned findings beyond those requested (\emph{over-release}), and $2$
omitted requested findings available in the chart (\emph{missed matches}).
The observed agreement with human review was therefore $95\%$.

\subsection{Re-grading of the Retrieval Comparison}
\label{app:sym-grading}

Human re-grading confirms the retrieval gains: an independent reviewer gives source-clean consultations higher \textsc{Accurate} rates than video consultations for all three re-graded systems, with the largest gains for MiMo-v2.5 and MiniMax-M3 (Table~\ref{tab:sym-grading}). For each clip of GPT-5.6-luna, MiMo-v2.5 and MiniMax-M3, the reviewer saw the confirmed diagnosis, the accepted and partial-credit lists, and the video and source-clean final diagnoses in random order without condition labels, and assigned each the three-level grade of Appendix~\ref{app:prompts}. The baseline is the video arm of Table~\ref{tab:consultation_conditions}, as in Figure~\ref{fig:lit}, and both graders of a system assessed the same consultation outputs.

\begin{table}[htbp]
\centering
\small
\caption{\textbf{Paired re-grading of video-condition and source-clean final diagnoses.}
\textsc{Accurate} rates in percent; $\Delta$ is source-clean minus video (pp);
paired clip counts follow each system.}
\label{tab:sym-grading}
\begin{tabular}{@{}llccc@{}}
\toprule
system & grader & video & source-clean & $\Delta$ \\
\midrule
GPT-5.6-luna ($71$) & human      & 46.5 & 53.5 & $+7.0$  \\
                    & automatic  & 46.5 & 49.3 & $+2.8$  \\
\addlinespace
MiMo-v2.5 ($71$)    & human & 26.8 & 47.9 & $+21.1$ \\
                    & automatic  & 22.5 & 43.7 & $+21.1$ \\
\addlinespace
MiniMax-M3 ($71$)   & human & 31.0 & 45.1 & $+14.1$ \\
                    & automatic  & 25.4 & 40.8 & $+15.5$ \\
\bottomrule
\end{tabular}
\end{table}

\end{document}